\documentclass[twoside,final,1p,times]{elsarticle}
\usepackage{amsmath}
\usepackage{multirow}
\usepackage{url}
\usepackage{fancyhdr}
\usepackage{color}
\usepackage{listings}
\journal{Data Intelligence}

\newcommand\shorttitle{\textcolor[RGB]{20,180,190}{Preprint submitted to Data Intelligence}}
\begin{document}
\begin{frontmatter}
\title{Exploring a Large Language Model for Transforming Taxonomic Data into OWL: Lessons Learned and Implications for Ontology Development}

\author[usp_poly,usp_ai,utwente]{Filipi Miranda Soares\corref{cor1}}
\author[usp_poly,usp_ai]{Antonio Mauro Saraiva}
\author[utwente]{Luís Ferreira Pires}
\author[utwente,lumc]{Luiz Olavo Bonino da Silva Santos}
\author[usp_math,usp_ai]{Dilvan de Abreu Moreira}
\author[usp_agro,usp_ai]{Fernando Elias Corrêa}
\author[usp_stats,usp_ai]{Kelly Rosa Braghetto}
\author[embrapa,usp_ai]{Debora Pignatari Drucker}
\author[usp_math,usp_ai]{Alexandre Cláudio Botazzo Delbem}

\cortext[cor1]{Corresponding author: Filipi Miranda Soares (f.mirandasoares@utwente.nl; ORCID: 0000-0002-0674-7960).}

\address[usp_poly]{University of Sao Paulo, Polytechnic School, Computer Engineering and Digital Systems, Sao Paulo, 05508-010, Brazil}
\address[usp_ai]{University of Sao Paulo, Center for Artificial Intelligence (C4AI), Sao Paulo, 05508-020, Brazil}
\address[utwente]{University of Twente, Faculty of Electrical Engineering, Mathematics and Computer Science, Semantics, Cybersecurity \& Services Group, Enschede, 7522 NB, Netherlands}
\address[lumc]{Leiden University Medical Center, Human Genetics, Leiden, 2333 ZC, Netherlands}
\address[usp_math]{University of Sao Paulo, Institute of Mathematics and Computer Science, Sao Carlos, 13566-590, Brazil}
\address[usp_agro]{University of Sao Paulo, Luiz de Queiroz College of Agriculture, Center for Advanced Studies on Applied Economics, Piracicaba, 13400-970, Brazil}
\address[usp_stats]{University of Sao Paulo, Institute of Mathematics and Statistics, Sao Paulo, 05508-090, Brazil}
\address[embrapa]{Embrapa Digital Agriculture, Campinas, 13083-886, Brazil}

\begin{abstract}
Managing scientific names in ontologies that represent species taxonomies is challenging due to the ever-evolving nature of these taxonomies. Manually maintaining these names becomes increasingly difficult when dealing with thousands of scientific names. To address this issue, this paper investigates the use of ChatGPT-4 to automate the development of the \texttt{\:Organism} module in the Agricultural Product Types Ontology (APTO) for species classification. Our methodology involved leveraging ChatGPT-4 to extract data from the GBIF Backbone API and generate OWL files for further integration in APTO. Two alternative approaches were explored: (1) issuing a series of prompts for ChatGPT-4 to execute tasks via the BrowserOP plugin and (2) directing ChatGPT-4 to design a Python algorithm to perform analogous tasks. Both approaches rely on a prompting method where we provide instructions, context, input data, and an output indicator. The first approach showed scalability limitations, while the second approach used the Python algorithm to overcome these challenges, but it struggled with typographical errors in data handling. This study highlights the potential of Large language models like ChatGPT-4 to streamline the management of species names in ontologies. Despite certain limitations, these tools offer promising advancements in automating taxonomy-related tasks and improving the efficiency of ontology development.
\end{abstract}


\end{frontmatter}

\section{Introduction}

Ontologies capture a well-defined and well-expressed shared conceptualization of a domain \cite{Yu2011}. They essentially specify the kinds of entities relevant to a particular domain and the relationships among such entities, thereby constructing a large semantic structure that represents the key aspects of domain knowledge \cite{sarika2014}. Ontologies also play a crucial role in facilitating the integration of heterogeneous data sources, enabling interoperability among information systems \cite{Vidal_2010}.

Ontology development is a comprehensive process involving the conceptualization step, during which domain experts and ontologists work together to specify significant concepts and relationships peculiar to a particular domain. Such collaboration guarantees that the established ontology sufficiently addresses the intricacies of the domain while being responsive to new knowledge and uses \cite{Femi2020,Campos_2009}. Ontologies facilitate the structured representation of knowledge, enhancing interoperability and reusability in modeling real-world concepts \cite{Ruiz_2006}.

Ontologies in the agricultural domain often incorporate taxonomic classifications to represent and organize knowledge about organisms with agricultural interest. Biologists define taxonomy as ``the theory and practice of identifying, describing, naming, and classifying organisms" \cite{Gillott_1995}. A taxonomy can be considered an ontology, as it formalizes hierarchical relationships and categories among organisms. However, the science of taxonomy faces numerous challenges from the dynamic nature of scientific discovery, as scientists continuously find new species and revise earlier established classifications \cite{Peruzzi_2023}. The current debate over competing classification systems, such as the Linnaean hierarchy versus phylogenetic taxonomy, further compounds this situation by not reaching a consensus \cite{Kuntner_Agnarsson_2006}. Additionally, maintaining consistency in naming conventions across languages and disciplines, managing synonyms (multiple names for the same organism), and resolving homonyms (the same name referring to different organisms) are also issues that have to be handled \cite{Rouhan_Gaudeul_2021}.

Despite these challenges, taxonomic classifications remain a valuable resource for structuring agricultural ontologies, aiding in the standardized classification of organisms of agricultural interest. Many approaches exist for integrating taxonomic data into these ontologies, ranging from manual to semi-automated methods. In manually curated vocabularies such as AGROVOC and Agrotermos, curators enter taxonomic names by completing forms to add new concepts, ensuring alignment with authoritative taxonomic databases \cite{AGROVOC_2023}. In contrast, semi-automated methods, such as those employed by the Food Ontology (FoodOn), leverage computational tools to streamline taxonomy integration. For example, FoodOn applies a ROBOT template-driven approach to refine seafood nomenclature, facilitating the alignment with NCBI GenBank names and NCBITaxon \cite{Sehar2021}.

Building on existing taxonomy integration approaches, this study evaluates the potential of large language models (LLMs) to convert taxonomic data into OWL format for integration into an agricultural ontology. Specifically, it leverages ChatGPT-4 to extract taxonomic data from the Global Biodiversity Information Facility (GBIF) Backbone Taxonomy API, which is then converted into OWL and integrated into the ontology. Two methods are explored: direct API querying via ChatGPT and a Python-based approach using LLM-generated code. The study assesses these methods in terms of scalability, reliability, and effectiveness in incorporating taxonomic data into a domain-specific ontology. In particular, it examines the feasibility of a GPT-driven approach for enhancing the Agricultural Product Types Ontology (APTO) by improving taxonomic accuracy and addressing ambiguities in agricultural product names, which often vary across languages, regions, and cultural contexts.

The paper is further organized as follows: Section \ref{sec:relatedWork} reviews related work, Section \ref{sec:apto} introduces APTO, Section \ref{sec:method} outlines the methods employed in this study, Sections \ref{approach1} and \ref{sec:approach2} present the results of Approach 1 and 2, respectively, Section \ref{sec:InteAPTO} explains the integration of species classification into APTO, and Section \ref{sec:final} concludes with the final considerations.

\section{Related Work}\label{sec:relatedWork}

The application of LLMs in Knowledge Graph (KG) and ontology development has gained attention due to their ability to process and generate conceptual schemas from natural language text sources \cite{feng2024ontology}. Recent studies have demonstrated that LLMs, such as ChatGPT, can significantly support tasks in ontology engineering, including concept extraction, class hierarchy generation, and semantic relationship identification \cite{feng2024ontology,Saeedizade_2024}.

A popular use of LLMs in ontology development is automating concept extraction and populating ontologies with domain knowledge. For example, Aggarwal et al. \cite{Aggarwal_2024} experimented with using LLMs in academic ontology development in engineering fields, with promising results in extracting and structuring domain knowledge. Zaitoun et al. \cite{Zaitoun_2023} also employed LLMs to enrich biomedical ontologies, demonstrating that they are able to identify missing concepts and relations. Trajanoska et al. \cite{Trajanoska_2023} and Reis et al. \cite{reis2024generating} demonstrated how LLMs can be utilized for enhancing KG construction from unstructured text.

LLMs also facilitate semantic relationship discovery and ontology alignment. Amini et al. \cite{Amini_2024} investigated LLMs for complex ontology alignment, using contextual embeddings to improve relationship detection. Matentzoglu et al. \cite{Matentzoglu_2023} demonstrated how LLMs like MapperGPT enhance entity linking and mapping precision. Wu et al. \cite{Wu_2024} suggested an approach using online clustering with LLMs for ontology expansion.

Additionally, studies have explored applying LLMs for entity resolution, focusing on identifying, matching, and reconciling entities across different data sources. Hees \cite{hees2024exploring} examined LLMs for aligning non-ontological data to ontologies, focusing on semantic interoperability in the energy sector. Using GPT-3.5 and GPT-4, the study aligned datasets from the ENERSHARE project with the Semantic TreeHouse ontology framework. The findings suggest that while LLMs support entity resolution, they function best as recommendation tools rather than deterministic alignment methods.

Zhang et al. \cite{Zhang2024Automated} explored weakly-supervised taxonomy enrichment, demonstrating how LLMs can identify discriminative features to enhance hierarchical structures. Their study introduced self-supervised entity resolution techniques and seed-guided taxonomy construction, showing that minimal human supervision enables LLMs to build structured ontologies from text corpora.

Hofer et al. \cite{Hofer2024Construction} analyzed LLM-driven entity reconciliation in KG construction, emphasizing the need for incremental ontology updates. Their findings indicate that while LLMs generate entity alignment candidates, a hybrid approach combining LLMs with rule-based validation is necessary for improving accuracy. The study highlights LLM applications in ontology alignment, particularly in biomedical KGs, where they refine taxonomic structures while maintaining semantic consistency.

Chen et al. \cite{Chen2023Knowledge} examined schema-based KGs for taxonomic reconciliation, leveraging OWL ontologies, RDFS, and SHACL. The study underscores the importance of ontology mapping repositories (e.g., UMLS Metathesaurus, BioPortal, MONDO) in resolving cross-domain taxonomic inconsistencies. Additionally, it explores LLM integration in entity resolution, particularly in instance matching and automated ontology extension.

These works emphasize the potential for LLMs to automate and improve ontology engineering, which is also investigated in this study. Table \ref{tab:related_studies} summarizes how these works relate to our study.

\begin{table}[ht]
\centering
\caption{Relationship between related studies and this paper}
\label{tab:related_studies}
{\scriptsize 
\renewcommand{\arraystretch}{0.9}
\begin{tabular}{p{3cm}p{4cm}p{5cm}}
\hline
\textbf{Study} & \textbf{Focus area} & \textbf{Relationship with this study} \\ \hline
Aggarwal et al. \cite{Aggarwal_2024} & LLMs for scholarly ontology generation & Automated concept and relationship extractions in ontology development \\
Zaitoun et al. \cite{Zaitoun_2023} & LLMs for biomedical ontology augmentation & Demonstrates LLMs' ability to identify missing ontological relationships \\
Trajanoska et al. \cite{Trajanoska_2023} \& Reis et al. \cite{reis2024generating} & LLMs for KG construction from unstructured text & LLM automation of ontology engineering tasks \\
Amini et al. \cite{Amini_2024} & LLMs for ontology alignment using contextual embeddings & Enhances taxonomic data alignment through contextual embeddings \\
Matentzoglu et al. \cite{Matentzoglu_2023} & LLMs for entity linking and mapping & Improves entity matching, crucial for taxonomic reconciliation \\
Wu et al. \cite{Wu_2024} & LLMs for ontology expansion using online clustering & Provides insights into LLM-based ontology expansion \\
Hees \cite{hees2024exploring} & LLMs for semantic interoperability and ontology alignment & Applies LLMs to ontology alignment, relevant for aligning non-ontological data to an ontology \\
Zhang et al. \cite{Zhang2024Automated} & LLMs for weakly-supervised taxonomy enrichment & Supports taxonomy enrichment with minimal human supervision \\
Hofer et al. \cite{Hofer2024Construction} & LLMs for incremental ontology updates and entity reconciliation & Enhancing the scalability of ontology completion, updates, and creation using LLMs to extract and integrate new concepts from structured data, such as tabular datasets. \\
Chen et al. \cite{Chen2023Knowledge} & KGs for taxonomic reconciliation and LLM integration & Demonstrates how schema-based KGs aid taxonomic integration with LLMs \\ \hline
\end{tabular}
}
\end{table}

Despite these developments, integrating LLMs into ontology engineering is challenging. Issues related to model interpretability and reliability \cite{Saeedizade_2024,tsaneva2024llm,perera2024exploring}, domain specificity, and computational costs \cite{Mulayim2024} are some of the key obstacles. Furthermore, the reliability of LLM-generated outputs in highly technical fields remains a concern, as hallucination and data bias continue to affect the consistency and accuracy of results \cite{Neuhaus_2023}. Overcoming these challenges is crucial to guaranteeing the robustness and real-world applicability of LLM-supported ontology engineering.

\section{The Agricultural Product Types Ontology (APTO)}\label{sec:apto}

APTO is a domain-specific ontology we developed to standardize terminology for agricultural product types within the Brazilian commodities market and comprises 3,333 statements. It incorporates over 200 product types identified in datasets from three Brazilian agricultural price index agencies, namely Cepea\footnote{URL: \url{https://www.cepea.esalq.usp.br/en}}, Ipea\footnote{URL: \url{https://www.ipeadata.gov.br/}}, and Conab\footnote{URL: \url{https://www.conab.gov.br/info-agro/precos}}. 

The ontology adopts a modular approach, comprising two core segments: the Organism module and the Product Type module. The Organism module links agricultural products to their biological origins using taxonomic classifications aligned with the GBIF Backbone Taxonomy, while the Product Type module categorizes products based on their origin (e.g., plant-based, animal-based, or inorganic) and processing state (raw, processed, or by-product). This duality allows APTO to capture the intricate relationships between products, their sources, and their processing stages, making it generally suitable for applications ranging from semantic web integration to agricultural market analysis.

We reused existing domain ontologies such as AGROVOC\footnote{URL: \url{https://agrovoc.fao.org/browse/agrovoc/en/}} \cite{SUBIRATSCOLL2022} and Agrotermos\footnote{URL: \url{https://sistemas.sede.embrapa.br/agrotermos/}} \cite{Telles2024Embrapa} when designing APTO. AGROVOC is a multilingual thesaurus developed by the Food and Agriculture Organization (FAO) to standardize terminology in agriculture, fisheries, forestry, and related domains \cite{SUBIRATSCOLL2022,Caracciolo2013}. In contrast, Agrotermos is a Brazilian agricultural vocabulary designed to support research and dissemination of knowledge specific to the Brazilian agricultural sector \cite{Pierozzi2020}. Additionally, Agrotermos serves as a terminology provider to AGROVOC for Brazilian Portuguese and incorporates concepts from AGROVOC into its vocabulary, ensuring interoperability and alignment between the two resources. 

\begin{figure}[ht]
    \centering
    \includegraphics[width=0.8\textwidth]{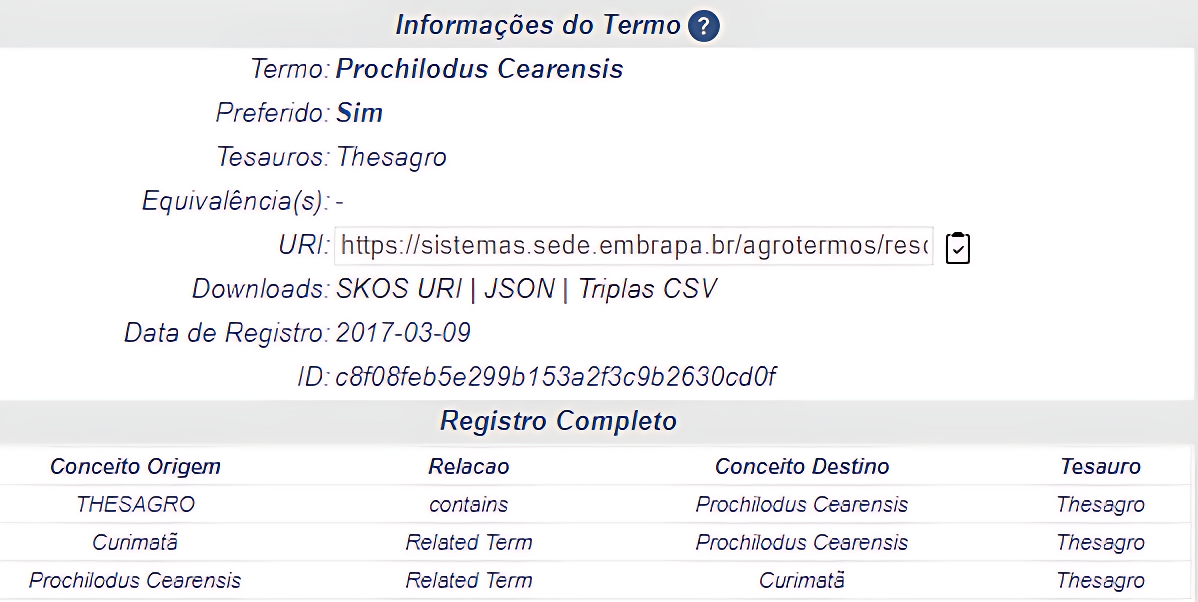}
    \caption{Example of synonym in Agrotermos in Portuguese\protect\footnotemark.}
    \label{fig:agrotermos}
\end{figure}

Despite their utility, a review of these vocabularies revealed gaps, particularly in the consistency and currency of taxonomic names. The dynamic nature of taxonomies, where scientific names are frequently updated due to synonymization and other taxonomic revisions, presents a significant challenge for maintaining ontological accuracy. For example, Agrotermos contains outdated taxonomic names due to the lack of automated mechanisms for verifying authoritative taxonomic sources for synonymization. Similarly, AGROVOC relies on a manual process for incorporating scientific names into its vocabulary as specified by its editorial guidelines \cite{AGROVOC_2023}, which can result in similar issues. Fig. \ref{fig:agrotermos} illustrates this problem with the species \textit{Prochilodus cearensis} Steindachner, 1911, which is a preferred term in Agrotermos, yet it is declared a synonym of \textit{Prochilodus brevis} Steindachner, 1875 in the GBIF Backbone Taxonomy \cite{prochilodus_cearensis_2023}.

\footnotetext{URL: \url{https://sistemas.sede.embrapa.br/agrotermos/resources/c8f08feb5e299b153a2f3c9b2630cd0f}}

We did not import species name classes directly from Agrotermos and AGROVOC to prevent these inconsistencies from propagating into APTO. Instead, we developed a method to extract up-to-date data from the GBIF Backbone Taxonomy and convert it into OWL format for seamless integration into the ontology.

The development of APTO followed the SABiO methodology for ontology development \cite{Falbo2014}, which encompasses five main phases: (1) Purpose Identification and Requirements Elicitation, (2) Ontology Capture and Formalization, (3) Design, (4) Implementation, and (5) Testing. This paper focuses on the Design phase, where we explored the use of ChatGPT to streamline the integration of taxonomic data into the ontology.

APTO has been implemented in OWL and is accessible through endpoints, including a SPARQL endpoint\footnote{URL: \url{https://data.pldn.nl/FilipiSoares/APTO/sparql}}, a text-search interface\footnote{URL: \url{https://data.pldn.nl/FilipiSoares/APTO/elasticsearch/APTO-Text-search/text}}, and an OWL/XML serialization\footnote{URL: \url{https://w3id.org/APTO\#}}. For Findability, Accessibility, Interoperability, and Reusability (FAIR) purposes, the ontology has been made available on AgroPortal\footnote{URL: \url{https://agroportal.lirmm.fr/ontologies/APTO}}.

\section{Methods}\label{sec:method}

Ontologies and KGs are closely related, with ontologies often serving as foundational frameworks for creating and interpreting KGs. Ontologies provide structured representations of domain-specific knowledge, including classes, relationships, and constraints. This logical foundation aligns with the principles of KGs, which are graph-based structures used to model interconnected data for reasoning and discovery \cite{kroetzsch2017ontologies,kroetzsch2016ontologies}. Ontologies not only aid in formalizing the semantics of KGs but also extend their capabilities by enabling advanced reasoning tasks \cite{kroetzsch2016ontologies}. Furthermore, they enhance KG construction by ensuring semantic interoperability and consistency, particularly when generating KGs from textual data \cite{Mihindukulasooriya2023}. The overlap between ontologies and KGs illustrates how these paradigms coalesce to support complex data representation and reasoning \cite{Hogan2021}. Given this close relationship, ontology development can benefit from recent techniques employed in KG construction, particularly applying LLMs.

The method employed in this case builds upon the framework outlined by \cite{Pan2024} for LLM-augmented KG development to explore this synergy. The authors address different techniques using LLMs in KG development, including LLM-augmented KG Construction, and describe two approaches: end-to-end KG construction and distilling KGs from LLMs. According to \cite{Pan2024}, we can understand these approaches as follows:

\begin{itemize}
    \item \textbf{End-to-end KG construction} is an innovative approach utilizing LLMs to streamline KG creation from raw text, bypassing the conventional multi-stage process \cite{Pan2024}. Studies like \cite{Kumar2020} have developed methods integrating LLMs for named entity recognition and relation extraction, employing dual BERT-based classifiers to identify relationships between entities \cite{Pan2024}. This method allows for the direct assembly of KGs, showcasing models like Grapher and PiVE that facilitate KG generation from textual descriptions or correct errors in existing KGs using iterative verification with smaller LLMs \cite{Pan2024}.
    
    \item \textbf{Distilling KGs from LLMs} capitalizes on the inherent knowledge encoded within LLMs to extract and structure this information into KGs \cite{Pan2024}. Techniques like COMET \cite{Bosselut_2019} leverage LLMs to generate commonsense KGs by training on seed sets of knowledge tuples, enabling the LLM to produce novel, high-quality knowledge tuples \cite{Pan2024}. This approach underscores the potential of LLMs as tools for enhancing KG construction and as vast repositories of implicit knowledge ripe for extraction and organization into explicitly structured forms \cite{Pan2024}.
\end{itemize}

The methodological approach employed in this paper is similar to the End-to-End KG Construction, but less complex since we do not deal with extraction from raw text. The data input we used in the ontology construction is structured as JSON files with representative and precise names, which makes it easier to process (and more precise) than natural language input. 

\begin{figure}[ht]
\centering
\includegraphics[width=\linewidth]{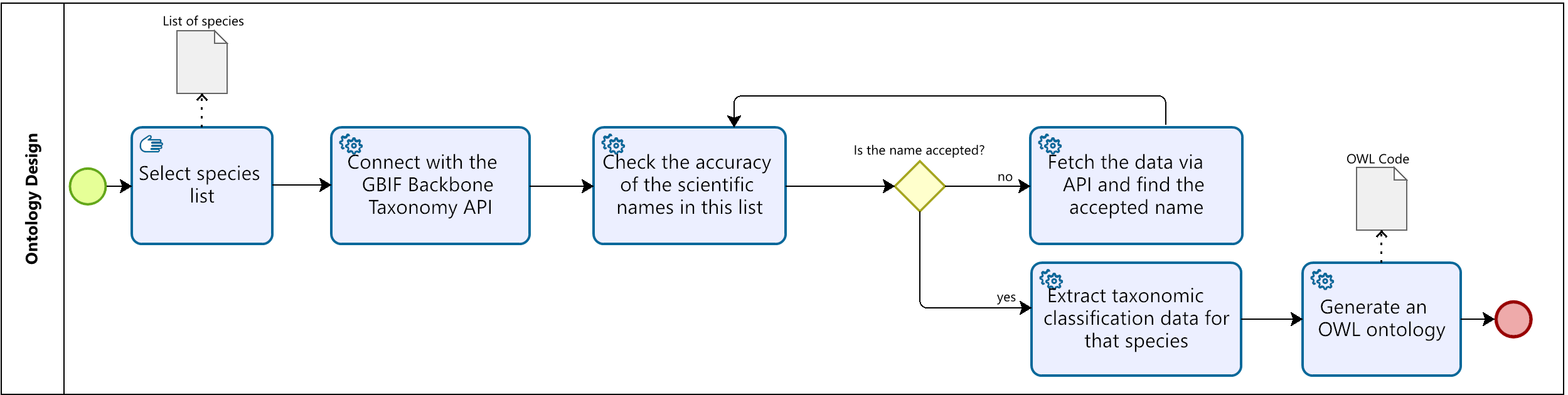}
\caption{Workflow for the taxonomic data transformation.}
\label{fig:pipeline}
\end{figure}

In this research, we used ChatGPT-4 to improve the quality of an ontology. Two alternative approaches were explored: (1) issuing a series of prompts for ChatGPT-4 to execute tasks via the BrowserOP plugin and (2) directing ChatGPT-4 to design a Python algorithm to perform analogous tasks. We followed the workflow activities shown in Fig. \ref{fig:pipeline} in both approaches. The following sections discuss each activity in detail. 

\subsection{Selecting the Species List}\label{sec:specieslist}

We started by selecting a list of plant and animal species related to all agricultural product types in APTO. This step involved mapping product types from three key Brazilian datasets onto commodity price indexes, namely the agricultural prices index from the Center for Advanced Studies on Applied Economics (Cepea) \cite{cepea_website}, the IpeaData Macroeconomic Indicators \cite{ipeadata_website}, and the 'Companhia Nacional de Abastecimento' (Conab) agricultural prices database \cite{conab_website}. We based the association between product types and species on resources such as Agrotermos, Agrovoc, and the scientific literature, producing two lists: an Animal species list with 14 scientific names and a Plant species list with 74 names, both lists with names of species or genus level.

Another criterion for selecting these species was their agricultural relevance, particularly within Brazilian agronomic and livestock domains. The list includes economically significant species, widely cultivated, and well-documented in key Brazilian agricultural data repositories. Given Brazil’s diverse agricultural landscapes, these species provide a representative yet incomplete view of the taxonomic groups involved in commercial agriculture.

While the selection is representative of Brazilian agrobiodiversity, it has limitations when generalizing to global taxonomies. Many agricultural taxonomies include species from other regions with different climatic, economic, and ecological contexts. As this study focuses on Brazilian datasets, certain species with regional importance elsewhere may be underrepresented. However, the approach developed in this study is not limited to Brazilian species, as it can be extended to include species from other regions, as long as they are represented in the GBIF Backbone Taxonomy. 

\subsection{Connecting with the GBIF Backbone Taxonomy API}

GBIF is an international network and data infrastructure supported by governments worldwide. Its primary goal is to provide open access to biodiversity data for all types of life on Earth, accessible from anywhere \cite{GBIF2024}. Recognized as the most extensive biodiversity data infrastructure regarding the volume of observations \cite{Sterner_2023}, GBIF serves many stakeholders, including researchers, policymakers, and other decision-makers. It achieves this by aggregating data from numerous sources, standardizing the information, and making it freely available for public use \cite{CBDGBIF2018}.

For the Organism module in APTO, we imported scientific names from the GBIF Backbone Taxonomy \cite{gbif2023backbone}. GBIF provides an API\footnote{URL: \url{https://api.gbif.org/v1/}} for extracting taxonomic classifications. This API operates on data stored within the GBIF Checklist Bank, which indexes every registered checklist dataset across the GBIF network \cite{gbifAPI}. To consume these RESTful JSON web services, GBIF internally utilizes a Java web service client \cite{gbifAPI}. Using ChatGPT-4 and the BrowserOp plugin, we automated the retrieval of taxonomic data for each species via the GBIF API.

The GBIF Backbone Taxonomy is a comprehensive synthetic classification system that underpins GBIF's integration of diverse name-based information from various sources, including occurrence datasets, species pages, and external databases like the Encyclopedia of Life (EOL), Genbank, and the International Union for Conservation of Nature and Natural Resources (IUCN) red list database, facilitating consistent taxonomic searches and cross-referencing across resources \cite{gbif2023backbone}. Regular updates extend its scope by merging additional scientific names from authoritative datasets into the Catalogue of Life-based structure \cite{Hobern_2021}, enhancing coverage beyond family-level classifications \cite{gbif2023backbone}. The taxonomy also incorporates identifiers for Operational Taxonomic Units (OTUs) from barcoding projects like iBOL and UNITE, employing consensus-based methods to connect Barcode Index Numbers (BINs) and Species Hypotheses (SHs) to taxon names and classifications \cite{gbif2023backbone}. Available for download in various formats, the GBIF Backbone Taxonomy data represents a critical resource for taxonomic research, drawing from over 105 sources to ensure broad and accurate name coverage \cite{gbif2023backbone}.

\subsection{Checking the Accuracy of Scientific Names}

International nomenclature codes govern the assignment of scientific names to species, establishing principles and rules to ensure clarity and consistency across the scientific community. This process involves the formal description and publication of each species, adhering to specific criteria set by these codes. Over time, scientific names may become deprecated due to various factors, such as the discovery of prior valid names overlooked in earlier assessments or new phylogenetic insights that necessitate reclassification \cite{Greuter1998,Winston2018}. As a result, the nomenclatural landscape is dynamic, with taxonomists revising and updating names to reflect the most accurate understanding of biological relationships. These updates lead to the synonymization of many species names \cite{Greuter1998,Winston2018}. 

We reviewed each name on the species list to determine whether it was accepted or a synonym. In Approach 1, we used ChatGPT to check the name status through the GBIF API. In Approach 2, we implemented this process using the Python Taxonomy Converter.

\subsection{Extracting Multi-level Taxonomic Classification Data}

Species classification follows various conventions, leading to multi-level taxonomies that organize life forms into hierarchical categories \cite{Okasha2019}. In this study, we adopted a seven-tiered framework for the taxonomic classification of species, consisting of kingdom, phylum, class, order, family, genus, and species, aligned with established biological taxonomy principles \cite{Okasha2019}. This hierarchical structure is crucial for understanding the relationships and distinctions among different life forms \cite{Okasha2019}. During the classification phase, we tasked both applications (ChatGPT integrated with the BrowserOp plugin and the Python Taxonomy Converter) with extracting these seven levels of classification for each species listed, or six levels for names at the genus level.

\subsection{Transforming Taxonomic Data Into OWL}\label{sec:owlgenerate}

Finally, we used the extracted taxonomic information to generate OWL/XML files, incorporating hierarchical relationships that accurately reflect the underlying species taxonomy upon which they were based.

\section{Approach 1: ChatGPT with BrowserOp Plugin}\label{approach1}
\subsection{Prompt Specification and Results}

The process consisted of incorporating a series of instructions into the ChatGPT 4 model, as shown in Table \ref{table:tasks}. The prompt engineering methodology used in this study builds on Giray's work \cite{Giray_2023}, which categorizes the components of an effective prompt into four main elements:

\begin{itemize}
    \item \textbf{Instruction}: A directive that specifies the task, guiding the model’s behavior and ensuring it aligns with the desired outcome.
    \item \textbf{Context}: Relevant background information or external details provide additional knowledge, enabling the model to generate accurate and context-aware responses.
    \item \textbf{Input data}: The core query or data to be processed serves as the primary driver of the model's understanding and response generation.
    \item \textbf{Output indicator}: An indicator of the expected response format, such as a brief answer, detailed explanation, or structured output, shaping the model's solution delivery.
\end{itemize}

We built the prompt based on these four elements, as shown in Table \ref{table:tasks}.

\begin{table}[ht]
\centering
\caption{Prompt Specification}
\label{table:tasks}
{\scriptsize 
\renewcommand{\arraystretch}{0.9} 
\begin{tabularx}{\textwidth}{>{\centering\arraybackslash}m{3.5cm}X}
\hline
\textbf{Element} & \textbf{Prompt} \\
\hline
Instruction & Build an ontology for species names. \\
Context & Access the GBIF API [using the BrowserOp plugin]: \url{https://api.gbif.org/v1/} \\
Input data & Extract the full taxonomic classification for each name in this list, including kingdom, phylum, class, order, family, genus, and species. In some cases, the name on the list is at the species level, and in other cases, at the genus level. \\
Input data & If a name has the status `deprecated`, look for the accepted name for that species. \\
Output indicator & This taxonomy should be organized as OWL classes and subclasses (do not create individuals or instances). Each class should have a URI from GBIF and an RDF label with the name for that class, with the language tag (considering that scientific names are written in Latin, include the language tag ``lat"). Finally, generate an OWL code, including all the necessary namespaces. \\
\hline
\end{tabularx}
} 
\end{table}

The species names used as \textbf{Input data} are listed in Table \ref{table:species}. The first species in this list, \textit{Apis mellifera}, served as a test case where the ChatGPT model utilized the BrowserOp plugin to send a request to the GBIF API. This plugin fetched real-time data from the web \cite{BrowserOP}, and facilitated the interaction as detailed in Listing \ref{lst:request_to_browserop}. 
As specified in \textbf{Context}, the call to the GBIF API performed with the BrowserOP plugin successfully retrieved the desired data. In our directive to ChatGPT, we specified a connection with the GBIF API's base URL (\url{https://api.gbif.org/v1/}), which hosts various endpoints catering to different data dimensions, including but not limited to the Registry API, Species API, Occurrence API, and Literature API, each serving distinct purposes such as registry lookup, species matching, occurrence data retrieval, and literature references, respectively. Despite the lack of explicit instruction on which endpoint to use, ChatGPT identified and targeted the appropriate API endpoint for species matching, which is \url{https://api.gbif.org/v1/species/}.

\begin{table}[ht]
\centering
\caption{List of animal scientific names}
\label{table:species}
{\scriptsize 
\renewcommand{\arraystretch}{0.9}
\begin{tabular}{p{3cm}p{4cm}}
\hline
\textbf{Taxon rank} & \textbf{Taxon name} \\ \hline
Species & \textit{Apis mellifera} \\ 
Species & \textit{Bos taurus} \\ 
Subspecies & \textit{Capra aegagrus hircus} \\ 
Species & \textit{Ovis aries} \\ 
Genus & \textit{Sus} \\ 
Species & \textit{Bombyx mori} \\ 
Species & \textit{Prochilodus cearensis} \\ 
Species & \textit{Prochilodus scrofa} \\ 
Species & \textit{Prochilodus margravii} \\ 
Species & \textit{Semaprochilodus insignis} \\ 
Species & \textit{Semaprochilodus taeniunes} \\ 
Species & \textit{Colossoma mitrei} \\ 
Species & \textit{Arapauma gigas} \\ 
Genus & \textit{Tilapia} \\ \hline
\end{tabular}
}
\end{table}

\FloatBarrier
\begin{lstlisting}[language=JSON, 
                   caption=Response from the GBIF API to the GPT-BrowserOP call, 
                   label=lst:request_to_browserop,
                   basicstyle=\footnotesize\ttfamily]
// Request to BrowserOP
{
  "link": "https://api.gbif.org/v1/species/match?name=Apis%20mellifera"
}

// Response from BrowserOp
{
  "result": {
    "content": "{\"usageKey\":1341976,\"scientificName\":\"Apis mellifera Linnaeus, 1758\",\"canonicalName\":\"Apis mellifera\",\"rank\":\"SPECIES\",\"status\":\"ACCEPTED\",\"confidence\":99,\"matchType\":\"EXACT\",\"kingdom\":\"Animalia\",\"phylum\":\"Arthropoda\",\"order\":\"Hymenoptera\",\"family\":\"Apidae\",\"genus\":\"Apis\",\"species\":\"Apis mellifera\",\"kingdomKey\":1,\"phylumKey\":54,\"classKey\":216,\"orderKey\":1457,\"familyKey\":4334,\"genusKey\":1334757,\"speciesKey\":1341976,\"synonym\":false,\"class\":\"Insecta\"}\n",
    "rules": [
      "Use the language that the user previously used or the language requested by the user.",
      "Based on the provided data (and all historical retrieved data if the results need more than 1 requests to be completely provided), respond to the user's request. No need to include other irrelevant words in your answer.",
      "Always extract url or source from provided content as a support for your answer.",
      "Ensure the information provided is accurate, clear, and concise. Avoid overly complex or lengthy explanations when a simple, direct response will suffice, while in the meantime, make sure to provide all the information the user will find valuable based on your correct understanding of user's intention.",
      "If a conclusive answer cannot be generated based on the provided content, respond to user's query first and then attempt to interpret the user's intent and suggest ways for them to refine their query for a more successful result.",
      "If the provided data includes authoritative source such as [wiki] or [official website], organize the response based primarily on data from these sources.",
      "If the user's query contains a URL, give a brief description of the site and summarize it based on the content of the page."
    ]
  },
  "code": 0
}
\end{lstlisting}

The API response shown in Listing~\ref{lst:request_to_browserop} adheres to the prompt specification (refer to Table \ref{table:tasks}), demonstrating the successful extraction of detailed taxonomic classification data for the species \textit{Apis mellifera}. As specified in the prompt's \textbf{Input data}, ChatGPT retrieved the complete taxonomic hierarchy, including the ranks kingdom, phylum, class, order, family, genus, and species. Listing \ref{lst:request_to_browserop} confirms the accurate extraction of all ranks, formatted as key-value pairs (e.g., \texttt{kingdom:Animalia}).

The prompt further specified that the ontology generated from the retrieved (\textbf{Output indicator} in Table~\ref{table:tasks}) data should incorporate GBIF URIs for each class representing a taxon. Listing~\ref{lst:request_to_browserop} shows that each taxon is assigned a unique \texttt{key} corresponding to its taxonomic rank (e.g., \texttt{kingdomKey:1}, \texttt{genusKey:133475}). GBIF assigns a unique numeric ID to each entity in its Backbone Taxonomy, forming the URI that uniquely identifies every scientific name in its catalog. For example, the response indicates Animalia is linked to the \texttt{kingdomKey} value 1, which is represented by the GBIF URI \url{https://www.gbif.org/species/1}. These URIs ensure that each taxon is unambiguously identified within the ontology.

The prompt also required ChatGPT to verify the status of the species' name. The API response in Listing~\ref{lst:request_to_browserop} shows the name status as \texttt{ACCEPTED}, indicating that the species name is valid and recognized within the taxonomy catalog. This outcome confirms ChatGPT's successful execution of this verification step.

Moreover, Listing~\ref{lst:request_to_browserop} shows the \texttt{rules} block, which outlines a set of operational guidelines not explicitly defined in the prompt, suggesting they originate from ChatGPT's model internal rules to enhance the response's comprehensiveness and relevance.

Finally, the prompt in Table~\ref{table:tasks} defined that the output should be an OWL file generated from the data extracted via the API. Initially, GPT-4 created OWL code using \url{https://api.gbif.org/v1/species/} as the namespace for the ontology classes. This URI is a reference point within the GBIF API, linking to JSON data files describing scientific names and their relationships, as illustrated in Listing \ref{lst:animalia_class}.

\begin{lstlisting}[language=JSON, caption=JSON data for the Animalia class (https://api.gbif.org/v1/species/1), label=lst:animalia_class, basicstyle=\footnotesize\ttfamily]
{
  "key": 1,
  "nubKey": 1,
  "nameKey": 130188353,
  "taxonID": "gbif:1",
  "sourceTaxonKey": 1,
  "kingdom": "Animalia",
  "kingdomKey": 1,
  "datasetKey": "d7dddbf4-2cf0-4f39-9b2a-bb099caae36c",
  "constituentKey": "d7dddbf4-2cf0-4f39-9b2a-bb099caae36c",
  "scientificName": "Animalia",
  "canonicalName": "Animalia",
  "vernacularName": "Animals",
  "authorship": "",
  "nameType": "SCIENTIFIC",
  "rank": "KINGDOM",
  "origin": "SOURCE",
  "taxonomicStatus": "ACCEPTED",
  "nomenclaturalStatus": [],
  "remarks": "",
  "numDescendants": 2981931,
  "lastCrawled": "2023-08-22T23:20:59.545+00:00",
  "lastInterpreted": "2023-08-22T22:11:51.237+00:00",
  "issues": []
}
\end{lstlisting}

However, after careful consideration, we decided to use URIs from the \url{https://www.gbif.org/species/} endpoint instead. We chose this endpoint because it provides access to valuable enriched resources for end users, such as taxon photos, occurrence statistics, and related information. Although we did not import these supplementary resources into the ontology, users can access them by clicking on the URL, aligning with linked data principles.

To apply this change, we directed ChatGPT to regenerate the OWL code using \url{https://www.gbif.org/species/} as the namespace. Listing \ref{lst:owl_taxonomic_hierarchy} shows the updated output.

\begin{lstlisting}[language=owl, caption=GPT-generated OWL Representation for Taxonomic Hierarchy, label=lst:owl_taxonomic_hierarchy, breaklines=true, basicstyle=\footnotesize\ttfamily]
<rdf:RDF xmlns:rdf="http://www.w3.org/1999/02/22-rdf-syntax-ns#"
         xmlns:rdfs="http://www.w3.org/2000/01/rdf-schema#"
         xmlns:owl="http://www.w3.org/2002/07/owl#">

    <!-- Kingdom Animalia -->
    <owl:Class rdf:about="https://www.gbif.org/species/1">
        <rdfs:label xml:lang="lat">Animalia</rdfs:label>
    </owl:Class>

    <!-- Phylum Arthropoda -->
    <owl:Class rdf:about="https://www.gbif.org/species/54">
        <rdfs:label xml:lang="lat">Arthropoda</rdfs:label>
        <rdfs:subClassOf rdf:resource="https://www.gbif.org/species/1"/>
    </owl:Class>

    <!-- Class Insecta -->
    <owl:Class rdf:about="https://www.gbif.org/species/216">
        <rdfs:label xml:lang="lat">Insecta</rdfs:label>
        <rdfs:subClassOf rdf:resource="https://www.gbif.org/species/54"/>
    </owl:Class>

    <!-- Order Hymenoptera -->
    <owl:Class rdf:about="https://www.gbif.org/species/1457">
        <rdfs:label xml:lang="lat">Hymenoptera</rdfs:label>
        <rdfs:subClassOf rdf:resource="https://www.gbif.org/species/216"/>
    </owl:Class>

    <!-- Family Apidae -->
    <owl:Class rdf:about="https://www.gbif.org/species/4334">
        <rdfs:label xml:lang="lat">Apidae</rdfs:label>
        <rdfs:subClassOf rdf:resource="https://www.gbif.org/species/1457"/>
    </owl:Class>

    <!-- Genus Apis -->
    <owl:Class rdf:about="https://www.gbif.org/species/1334757">
        <rdfs:label xml:lang="lat">Apis</rdfs:label>
        <rdfs:subClassOf rdf:resource="https://www.gbif.org/species/4334"/>
    </owl:Class>

    <!-- Species Apis mellifera -->
    <owl:Class rdf:about="https://www.gbif.org/species/1341976">
        <rdfs:label xml:lang="lat">Apis mellifera</rdfs:label>
        <rdfs:subClassOf rdf:resource="https://www.gbif.org/species/1334757"/>
    </owl:Class>

</rdf:RDF>
\end{lstlisting}

We repeated this process for all taxa listed in Table \ref{table:species}. ChatGPT confirmed the taxon names \textit{Bos taurus}, \textit{Ovis aries}, Genus \textit{Sus}, \textit{Semaprochilodus insignis}, \textit{Semaprochilodus taeniurus}, \textit{Arapaima gigas}, and \textit{Tilapia} as accepted names within the GBIF Backbone Taxonomy. Then, it successfully generated the corresponding OWL code for these taxa. The OWL files are available on Zenodo [\ref{code:12684941}].

ChatGPT identified the remaining taxa as having outdated or deprecated names, classifying them as synonyms, as shown in Table \ref{table:deprecated}. These taxon names were listed as preferred terms in the Agrotermos thesaurus but were absent from AGROVOC, the two vocabularies used as the foundation for building this ontology. An exception is \textit{Capra hircus}, which we did not find in either of these vocabularies.

\begin{table}[ht]
\centering
\caption{Synonymous scientific names and their corresponding accepted names identified by ChatGPT}
\label{table:deprecated}
{\scriptsize 
\renewcommand{\arraystretch}{0.9} 
\begin{tabular}{p{4cm}p{4cm}}
\hline
\textbf{Target name} & \textbf{Synonym of} \\ \hline
\textit{Capra hircus} & \textit{Capra aegagrus} \\ 
\textit{Prochilodus cearensis} & \textit{Prochilodus brevis} \\ 
\textit{Prochilodus scrofa} & \textit{Prochilodus lineatus} \\ 
\textit{Prochilodus margravii} & \textit{Prochilodus argenteus} \\ 
\textit{Colossoma mitrei} & \textit{Piaractus mesopotamicus} \\ \hline
\end{tabular}
}
\end{table}

Agrotermos and AGROVOC include concepts for representing `goat', but with limitations. In Agrotermos, the term `Cabra' (Portuguese for goat) is not explicitly associated with its scientific species. In AGROVOC, \textit{Capra hircus} is modeled as an entry term for this concept. To address this gap, we referred to the classification provided on Wikipedia, which defines `goat' species as \textit{Capra hircus}. We added this name to the list, which turned out to be outdated. 

ChatGPT retrieved the corresponding accepted name from the GBIF Backbone Taxonomy for each synonym identified and generated OWL code incorporating these validated names, as shown in Listing \ref{lst:owl_synonym}.

\begin{lstlisting}[language=owl, caption=Output After Resolving Synonyms, label=lst:owl_synonym, breaklines=true, basicstyle=\footnotesize\ttfamily]
<rdf:RDF xmlns:rdf="http://www.w3.org/1999/02/22-rdf-syntax-ns#"
         xmlns:rdfs="http://www.w3.org/2000/01/rdf-schema#"
         xmlns:owl="http://www.w3.org/2002/07/owl#">

                                ...

    <!-- Species Prochilodus brevis (for Prochilodus cearensis) -->
    <owl:Class rdf:about="https://www.gbif.org/species/2352151">
        <rdfs:label xml:lang="lat">Prochilodus brevis</rdfs:label>
        <rdfs:subClassOf rdf:resource="https://www.gbif.org/species/2352148"/>
    </owl:Class>

    <!-- Species Prochilodus lineatus (for Prochilodus scrofa) -->
    <owl:Class rdf:about="https://www.gbif.org/species/2352154">
        <rdfs:label xml:lang="lat">Prochilodus lineatus</rdfs:label>
        <rdfs:subClassOf rdf:resource="https://www.gbif.org/species/2352148"/>
    </owl:Class>

    <!-- Species Prochilodus argenteus (for Prochilodus margravii) -->
    <owl:Class rdf:about="https://www.gbif.org/species/2352177">
        <rdfs:label xml:lang="lat">Prochilodus argenteus</rdfs:label>
        <rdfs:subClassOf rdf:resource="https://www.gbif.org/species/2352148"/>
    </owl:Class>

</rdf:RDF>
\end{lstlisting}

In Listing \ref{lst:owl_synonym}, we tasked ChatGPT with generating OWL code for three species within the same genus: \textit{Prochilodus Cearensis}, \textit{Prochilodus Scrofa}, and \textit{Prochilodus Margravii}. However, during our first attempt to extract data from the GBIF API using these names, ChatGPT reported that no data was available for these taxa. We discovered that this issue stemmed from formatting errors in the species names. According to international standards for species nomenclature \cite{ICZN2012,Turland2018ShenzhenCode}, species names should consist of a capitalized genus name followed by a lowercase specific epithet, often including the naming authority and year of identification. The names we used in the initial query, copied directly from the Agrotermos vocabulary, had capitalization errors in the specific epithets (e.g., \textit{Cearensis} instead of \textit{cearensis}), which rendered them unrecognizable to the GBIF API.

After correcting the formatting to align with these standards, we reissued the query, and ChatGPT successfully retrieved the taxonomic data. Table \ref{table:interaction_chatgpt} provides a detailed log of this interaction, highlighting the issue and the resolution process.

\begin{table}[ht!]
\centering
\caption{Iteration with ChatGPT}
\label{table:interaction_chatgpt}
{\scriptsize 
\renewcommand{\arraystretch}{0.9} 
\begin{tabularx}{\textwidth}{>{\centering\arraybackslash}m{1cm}X}
\hline
\textbf{Agent} & \textbf{Prompt} \\
\hline
\raisebox{-.5\height}{\includegraphics[width=0.6cm]{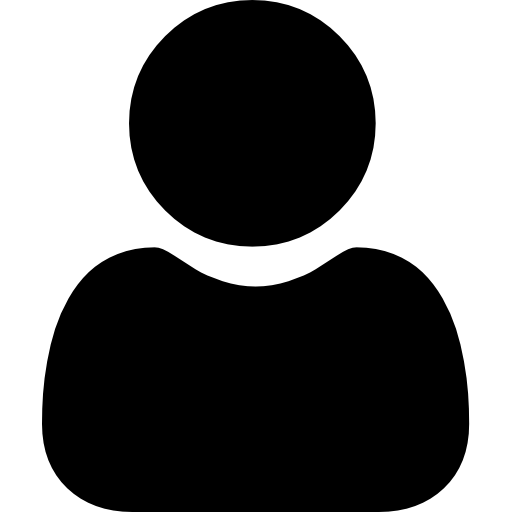}} & Do it [repeat the steps described in Table \ref{table:tasks}] for these three species: Prochilodus Cearensis, Prochilodus Scrofa, and Prochilodus Margravii. \\
\hline
\raisebox{-.5\height}{\includegraphics[width=0.6cm]{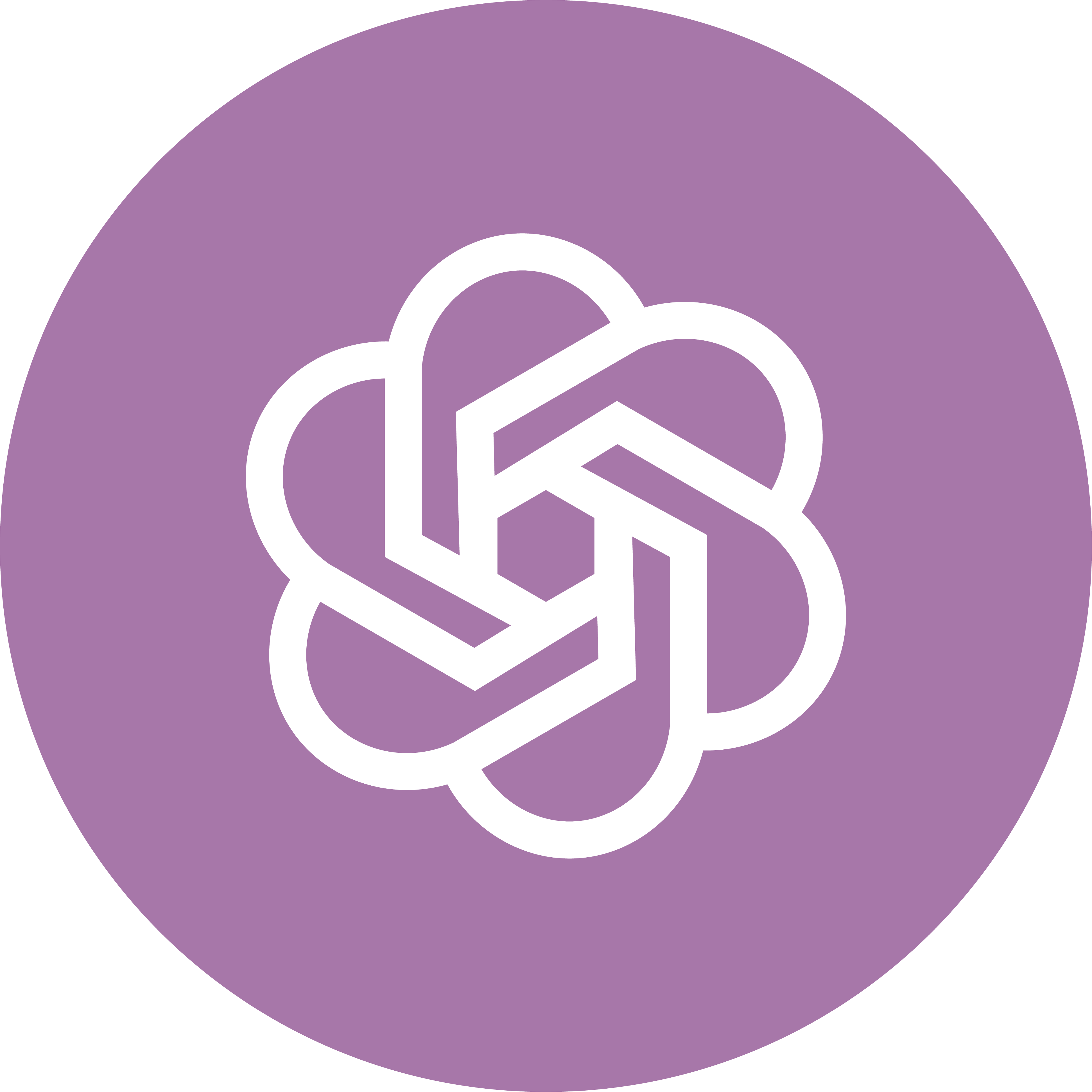}} & The attempts to match the species names \textit{Prochilodus Cearensis}, \textit{Prochilodus Scrofa}, and \textit{Prochilodus Margravii} with the GBIF database did not yield any results due to too little confidence in the matches. This means that the specific names provided might not be directly available in the GBIF database under those exact spellings, or are not recognized at the species level in the database.

For taxonomic research and ontology building, accurate species names are crucial. It is possible that these names might have synonyms under which they are listed in databases, or there might be slight variations in spelling or taxonomy that affect the search results.

In such cases, further steps could include:

Verifying the scientific names for spelling or taxonomic updates.
Searching for broader taxonomic groups (e.g., the genus ``Prochilodus") and then narrowing down to the species level based on additional sources or databases.
Consulting specialized taxonomic databases or literature for the most current classification and synonyms.
If you have alternative or broader taxonomic categories in mind or need assistance with another query, please let me know how I can assist further. \\
\hline
\raisebox{-.5\height}{\includegraphics[width=0.6cm]{Fig/user-icon.png}} & Try again with Prochilodus cearensis, Prochilodus scrofa, and Prochilodus margravii. \\
\hline
\raisebox{-.5\height}{\includegraphics[width=0.6cm]{Fig/chatgpt-logo.png}} & Based on the data retrieved from the GBIF API for the species Prochilodus cearensis, Prochilodus scrofa, and Prochilodus margravii, here's the ontology in OWL format. It's important to note that all three names are recognized as synonyms for other species within the GBIF database: [provided the OWL code shown in Listing \ref{lst:owl_synonym}. After the code block, it commented:] This OWL code represents the full taxonomic classification for the species Prochilodus cearensis (synonym for Prochilodus brevis), Prochilodus scrofa (synonym for Prochilodus lineatus), and Prochilodus margravii (synonym for Prochilodus argenteus), including the higher taxon levels (kingdom, phylum, class, order, family, genus) with each class having a valid URI from GBIF and a rdf label of the name for that class in Latin. \\
\hline
\end{tabularx}
}
\end{table}

Following the initial incident, we intentionally repeated the same mistake by submitting the species name \textit{Colossoma Mitrei} with the same capitalization error in the specific epithet. ChatGPT corrected the error autonomously before sending the request to the GBIF API, demonstrating its ability to learn iteratively from prior interactions. 

This adaptability reflects ChatGPT's advanced learning capabilities, allowing it to refine its comprehension and response strategies over time. Such functionality exemplifies a core characteristic of Artificial Intelligence (AI): the ability to internalize lessons from previous interactions and apply this knowledge to similar situations in the future, thereby reducing errors and enhancing operational efficiency \cite{wang_2023,Nagarhalli_2022}.

In ontology development, this AI-driven error correction potentially reduces manual effort and enhances data quality. Studies \cite{wang_2023,Nagarhalli_2022,Shaik_2022,Bulut_2022} highlight the growing role of AI in fostering intelligent and efficient system interactions, making it a transformative tool in fields that demand adaptability and precision.

\subsection{Limitations}

This approach presented limitations when handling extensive species lists. For this reason, we did not move forward with generating the OWL code for the plant species listed in the ontology since the list of plant species is larger than the animal species list. 

Moreover, the processing time increases with the list's size, impacting this method's scalability. For example, the processing time to call the GBIF API and generate the ontology for the species \textit{Apis mellifera} was 37 seconds. For the three species \textit{Prochilodus cearensis}, \textit{Prochilodus scrofa}, and \textit{Prochilodus margravi}, which were synonyms, the processing time was 2 minutes and 5 seconds. This approach proved efficient for smaller clusters of species names, particularly when the names are taxonomically related (within the same genus or family). However, we observed that reliability decreases for lists with more than five names, where the likelihood of hallucinations increases. Initially, we attempted to process the entire list of species names from Table~\ref{table:species} through ChatGPT in a single batch. However, this approach resulted in multiple inaccuracies, including the generation of URIs non-existent in the GBIF database, incorrect association between valid URIs and labels, and occasional omissions of tasks, such as adding RDF labels. We opted for a more controlled approach to address these issues by submitting names to ChatGPT individually or in smaller groups (at most three species each time). This strategy significantly improved the accuracy of the results.

\begin{figure}[ht]
\centering
\includegraphics[width=\linewidth]{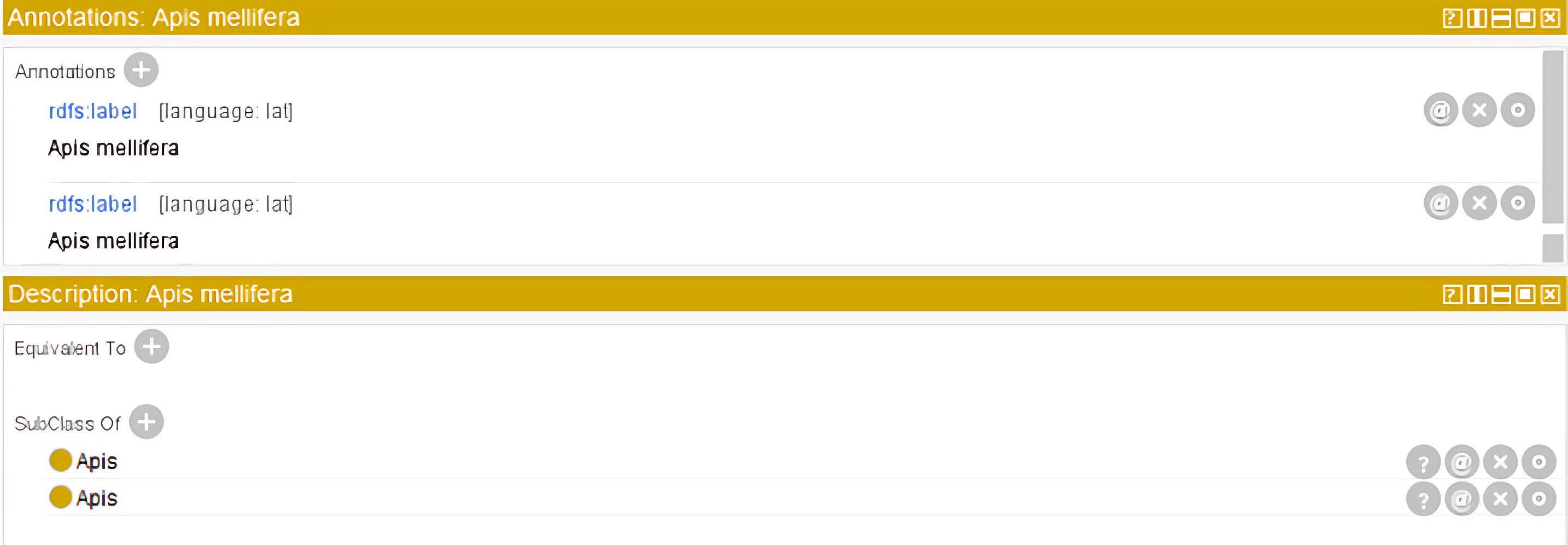}
\caption{Duplicated labels and relationships in the ontology visualization in Protégé.}
\label{fig:protege}
\end{figure}

Another challenge was integrating multiple OWL files using Protégé, which resulted in duplicated labels and parent-child relationships as illustrated in Fig. \ref{fig:protege}. Despite attempts to consolidate these files using Protégé's merge function, these redundancies persisted, but this can be also considered as a limitation of Protégé for handling these multiple files rather than a problem with ChatGPT-generated code. 

Another limitation of this approach is its poor reproducibility due to ongoing updates in ChatGPT’s technology. During the initial experiments, ChatGPT plugins such as BrowserOP were still operational. However, as of March 2024, OpenAI replaced these plugins with GPTs, and ChatGPT no longer has direct access to the Internet.

To assess the feasibility of reproducing the results, we attempted to replicate the experiments on March 6, 2025, using three different GPT models: BrowserOP (the former plugin), ChatGPT-4o, and Data Analyst. In all three cases, the models returned the same message indicating their inability to make direct API calls, confirming that ChatGPT currently lacks Internet access. Therefore, this approach is no longer viable unless ChatGPT restores internet access in the future.

To evaluate whether a similar approach could work with alternative AI models, we tested the same prompt using DeepSeek V3 and Google Gemini 2.0. However, the results were highly inaccurate and unreliable. Although both models generated API responses that structurally resembled the expected format (similar to Listing \ref{lst:request_to_browserop}), the responses contained incorrect taxon keys that did not match the expected labels. Additionally, neither model correctly identified that the species in this test (\textit{Prochilodus scrofa}) was a synonym.

Even after multiple refinements to the prompt, including providing more details about API syntax and sample responses, both models continued to generate hallucinated results that did not accurately reflect the expected outputs. These issues indicate that using this approach with DeepSeek V3 and Gemini 2.0 would require further fine-tuning or prompt engineering to mitigate these errors. The full prompts and interaction transcripts with DeepSeek and Google Gemini are available on Zenodo [\ref{code:12684941}].

In conclusion, this approach is currently not reproducible with ChatGPT due to the lack of internet access. Additionally, attempts to apply this methodology to other AI models, such as DeepSeek V3 and Gemini 2.0, require further adjustments to reduce hallucinations and improve reliability. Therefore, we recommend using Approach 2 as a more viable solution for generating an OWL ontology for species taxonomy.

\section{Approach 2: Python Algorithm for Converting Taxonomic Data to OWL}\label{sec:approach2}

Building on the steps outlined in Table \ref{table:tasks} and the prompt interactions described in Section~\ref{approach1}, we tasked ChatGPT with developing a Python algorithm to generate OWL files from data retrieved via the GBIF Backbone Taxonomy API. However, the initial version of the algorithm introduced redundancies in the generated OWL code by repeating higher taxonomic levels for each taxon. For example, the class \textit{Animalia} was redefined for every taxon describing an animal species rather than being defined once and used as a superclass for the relevant species. To address this issue, we provided ChatGPT with the revised instruction outlined in Table \ref{table:gpt1}. 

\begin{table}[ht]
\centering
\caption{Iteration with ChatGPT: fixing repetitions}
\label{table:gpt1}
{\scriptsize 
\renewcommand{\arraystretch}{0.9} 
\begin{tabularx}{\textwidth}{>{\centering\arraybackslash}m{1cm}X}
\hline
\textbf{Agent} & \textbf{Prompt} \\
\hline
\raisebox{-.5\height}{\includegraphics[width=0.6cm]{Fig/user-icon.png}} & It should present the superior taxon levels once in the code when they are the same for multiple species. Animalia, for instance, is a taxon common to all species in this list. \\
\hline
\raisebox{-.5\height}{\includegraphics[width=0.6cm]{Fig/chatgpt-logo.png}} & To address the issue of repeating higher taxon levels for multiple species, you can modify the script to accumulate unique taxon levels across all species before generating the OWL representation. This approach ensures that each taxon level (e.g., kingdom, phylum) is represented only once, regardless of how many species share it. \\
\hline
\end{tabularx}
}
\end{table}

The  algorithm second version fixed redundant taxonomic level definitions but introduced a new issue: improper class-subclass relationships in the OWL code. When reported to ChatGPT, it responded as shown in Table \ref{table:gpt2}.

\begin{table}[ht]
\centering
\caption{Iteration with ChatGPT: fixing relationships}
\label{table:gpt2}
{\scriptsize 
\renewcommand{\arraystretch}{0.9} 
\begin{tabularx}{\textwidth}{>{\centering\arraybackslash}m{1cm}X}
\hline
\textbf{Agent} & \textbf{Prompt} \\
\hline
\raisebox{-.5\height}{\includegraphics[width=0.6cm]{Fig/chatgpt-logo.png}} & ``To ensure the OWL output correctly represents the \texttt{subClassOf} relationships and avoids duplicating common higher taxon levels across species, let's refine the approach. We'll adjust the script to build a hierarchy that maintains these relationships and ensures each taxon level is included only once." \\
\hline
\end{tabularx}
}
\end{table}

In this third trial, we got the error: ``AttributeError: `list' object has no attribute `replace,' which was caused by a typo. We tasked ChatGPT to review the species list, and it detected \textit{Semaprochilodus taeniunes} as a typo, suggesting the correct name as \textit{Semaprochilodus taeniurus}. ChatGPT inference was correct since this is also the accepted name in the GBIF Backbone Taxonomy. After reviewing this name and reentering the list, the algorithm worked as expected. 

Finally, we tasked ChatGPT with generating the documentation for the algorithm, which we named Taxonomy OWLizer. The complete algorithm and its documentation are available on Zenodo [\ref{code:13328562}].

\subsection{Script Overview}

This Python script follows a pipeline to fetch taxonomic classification data for a species list from the GBIF API and generates an OWL representation. It can handle multiple species names, representing higher taxonomic levels only once, even if they are shared across species, avoiding duplication of classes as reported in the limitations of Approach 1. It also checks synonyms and looks for the accepted names. The script is written in Python 3.x and requires the \texttt{requests} library. 

\subsubsection{Functions}

This script met the requirements specified in Table \ref{table:tasks}, including the following functions: 

\begin{itemize}[nosep]
    \item \texttt{fetch\_gbif\_data(scientific\_name)}: Fetches taxonomic classification from the GBIF API for a given scientific name.
    \item \texttt{fetch\_synonyms(species\_key)}: Fetches synonyms for a given species key from the GBIF API. It checks the accepted scientific name of a species, ensuring that the OWL ontology only includes the correct names.
    \item \texttt{accumulate\_taxa(data, taxa)}: Accumulates unique taxa and their relationships from the GBIF data. 
    \item \texttt{validate\_accepted\_name(data, taxa)}: Ensures that only the accepted scientific name is included in the taxa dictionary by validating against synonyms. 
    \item \texttt{generate\_owl(taxa)}: Generates an OWL/XML representation from the accumulated taxa.
\end{itemize}
\medskip

\begin{figure}[ht!]
\centering
\includegraphics[scale=0.3]{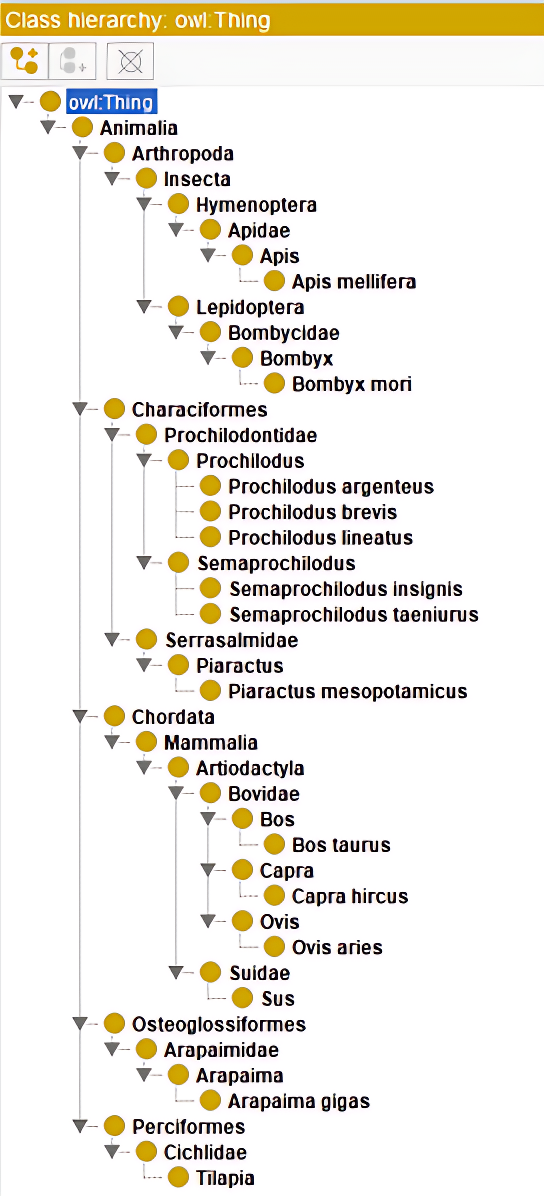}
\caption{Classes hierarchy in Protégé.}
\label{fig:taxo_prot}
\end{figure}

Listing \ref{lst:pipe} shows an example of usage for the list of names from Table \ref{table:species}. The output was an OWL/XML file (\texttt{Animalia.xml} on Zenodo [\ref{code:12684941}]). Subsequent inspection of this file in Protégé revealed a well-structured class hierarchy, depicted in Fig.~\ref{fig:taxo_prot}. This visualization confirmed the accuracy of the hierarchical relationships and the appropriate formatting of the generated OWL code.

\begin{lstlisting}[language=CustomPython, caption=Example of usage, label=lst:pipe, basicstyle=\footnotesize\ttfamily]
species_names = [
    "Apis mellifera", "Bos taurus", "Capra aegagrus hircus", "Ovis aries", "Sus", "Bombyx mori", "Prochilodus cearensis", "Prochilodus scrofa", "Prochilodus margravii", "Semaprochilodus insignis", "Semaprochilodus taeniurus", "Colossoma mitrei", "Arapaima gigas", "Tilapia"
]

taxa = {}
for name in species_names:
    data = fetch_gbif_data(name)
    if data:
        accumulate_taxa(data, taxa)
        validate_accepted_name(data, taxa)

owl_data = generate_owl(taxa)
print(owl_data)
\end{lstlisting}

\subsubsection{Testing the Algorithm with a Larger List of Species Names}

The scalability limitation of Approach 1 prevented the generation of OWL codes for long species lists. In contrast, the second approach works significantly faster. To evaluate its efficiency, we conducted a test with 74 plant species listed in the file \texttt{PlantSpeciesList.py} available on Zenodo [\ref{code:12684941}]. As an initial step, we asked ChatGPT to verify the spelling of the species names in this list using only its internal knowledge, without connecting to the GBIF API.

Although we did not explicitly request ChatGPT to identify synonyms since our algorithm handles this functionality, it flagged \textit{Euterpe edulis} as a synonym of \textit{Euterpe oleracea}. However, subsequent verification using authoritative sources, including the GBIF Backbone Taxonomy \cite{gbif2023backbone} and the Plants of the World Online catalog \cite{POWO2024}, confirmed that both \textit{Euterpe edulis} and \textit{Euterpe oleracea} are valid and accepted species names. This misidentification highlights the potential for inaccuracies when ChatGPT’s analysis is based solely on its internal knowledge rather than being anchored to reliable external sources for validating taxonomic names.

Despite the incorrect synonym suggestion, we continued with the task. The algorithm performed efficiently, generating OWL code for nearly all the listed species (except for \textit{Triticum x Secale}) in just 2 minutes and 31 seconds. By comparison, the first approach required 2 minutes and 5 seconds to generate OWL code for only three species. This stark contrast highlights the superior scalability of the second approach, which successfully processed 74 species at a similar time. The complete OWL code can be found as \texttt{plantae.xml} on Zenodo [\ref{code:12684941}].

\subsubsection{Hybrids in Ontological Modeling}

The exclusion of \textit{Triticum × Secale} from the output was due to inconsistencies in hybrid name formatting, which can vary across taxonomic catalogs. Hybrid species names are structured to indicate their origin from two different species or genera. While many taxonomic authorities use the symbol × to denote hybrid status -- following conventions in botanical and zoological nomenclature -- others may omit it \cite{ICZN2012,ICBNx}. For interspecific hybrids, the × is placed between the genus name and the specific epithet; for intergeneric hybrids, it appears between the genus names \cite{ICBNx}. For example, the intergeneric hybrid between Triticum (wheat) and Secale (rye) is typically written as \textit{Triticum × Secale}.

In the GBIF Backbone Taxonomy -- which aggregates names from various taxonomic sources -- the use of the hybrid symbol is inconsistent. For instance, \textit{Triticum secale}\footnote{URL: \url{https://www.gbif.org/species/4154790}} appears without the ×, while \textit{Citrus ×aurantium}\footnote{URL: \url{https://www.gbif.org/species/8077391}} includes the symbol but omits the space. Notably, the GBIF API supports both formats when queried. We tested the algorithm using both \textit{Triticum ×secale} (with the symbol and lowercase epithet) and \textit{Triticum secale} (without the symbol), and both returned results, indicating that the name is recognized as a synonym of \textit{Secale cereale}. To avoid issues during processing, hybrid names should be preprocessed into one of these formats before being submitted to the algorithm.

To enhance the expressiveness of hybrid modeling in our ontology, we manualy modeled hybrid relationships explicitly using an object property is\_a\_hybrid\_of. For instance, the triple:

\begin{verbatim}
Citrus aurantium is_a_hybrid_of some (Citrus maxima and Citrus reticulata)    
\end{verbatim}

captures the hybrid lineage of \textit{Citrus aurantium}. A corresponding OWL representation is shown in Listing \ref{lst:hybrids_owl}, where the IRI https://www.gbif.org/species/8077391 identifies the class \textit{Citrus aurantium}.

\begin{lstlisting}[language=owl, caption=Modeling of hybrids in APTO, label=lst:hybrids_owl]
        <Class IRI="https://www.gbif.org/species/8077391"/>
        <ObjectSomeValuesFrom>
            <ObjectProperty IRI="is_a_hybrid_of"/>
            <ObjectIntersectionOf>
                <Class IRI="Citrus_maxima"/>
                <Class IRI="Citrus_reticulata"/>
            </ObjectIntersectionOf>
        </ObjectSomeValuesFrom>
\end{lstlisting}

The is\_a\_hybrid\_of object property formally captures the hybridization relationship, linking the hybrid species class to the classes representing its parent species. This approach provides a precise and machine-interpretable way to represent genetic lineage within the ontology.

\subsection{Web Application with User Interface}

Building on the Python script of Approach 2, we developed the Taxonomy OWLizer app\footnote{URL: \url{https://filipi-soares.github.io/TaxonOWLConverter/}}, a web-based application that allows users to convert species names into an OWL taxonomy using the GBIF API. It provides a simple interface for users to enter species names, fetch their taxonomic classification, and download the generated OWL file. This tool is designed for biologists, taxonomists, and ontology researchers needing structured taxonomy data in OWL format.

The application supports the following key features:

\begin{enumerate}
    \item Users can input the names of one or multiple species.
    \item The app queries the GBIF API to fetch taxonomic classifications.
    \item The app converts the taxonomy data into an OWL ontology format.
    \item Users can copy and paste or download the generated OWL taxonomy file.
    \item Previously searched species are cached in local storage, allowing instant retrieval without re-fetching data.
\end{enumerate}

To develop the OWLizer, we utilized the following technologies:

\begin{itemize}
    \item HTML, CSS (Bootstrap): Provides a simple and responsive user interface.
    \item JavaScript: Handles API requests, caching, and file downloads.
    \item jQuery: Simplifies DOM manipulation and AJAX requests.
    \item Flask (Python Back-end on Render): Processes API requests and fetches taxonomic data from GBIF.
    \item GBIF API: Provides taxonomic classification data.
    \item GitHub Pages: Hosts the front-end (static website) for free.
    \item Render (Free Tier): Hosts the back-end Flask API that fetches data.
    \item UptimeRobot: Sends periodic requests to prevent the back-end from going to sleep.
\end{itemize}

\subsection{Limitations}

Approach 2 successfully addressed the scalability limitations of Approach 1. However, this algorithm remains sensitive to typographical errors in the input data. To mitigate this, we recommend a preliminary step in which users utilize ChatGPT to review the list of species names and correct typos. This workflow combines the strengths of both tools: ChatGPT's ability to verify and correct spellings, and the algorithm's efficiency in processing species names and performing synonym checks.

While the Taxonomy OWLizer web-based app provides a functional and accessible way to generate OWL representations of taxonomic data, it has certain limitations due to free-tier services. The back-end, which is hosted on Render’s free tier, enters a sleep state after 15 minutes of inactivity, causing delays in API responses when it needs to restart. To mitigate this, UptimeRobot periodically pings the back-end, keeping it active most of the time. However, this workaround is not entirely reliable and may still result in occasional downtime. Since GitHub Pages is used to host the front-end, the application can only serve static content, requiring an external back-end to handle API requests. 

Furthermore, the app caches responses using local storage, meaning that previously fetched data is only available on the same device and browser. This limits persistent caching across multiple sessions or devices, which could be improved by integrating a more robust storage solution such as IndexedDB or a cloud-based database. Despite these limitations, the current implementation remains an effective and free-to-use tool for taxonomists and ontology researchers.

\section{Integrating Species Classification in APTO}\label{sec:InteAPTO}

After generating and integrating the OWL file with species classifications into APTO Organism module, each product type was linked to its corresponding species. This association is essential for accurately representing the meaning of commercial product names, particularly within the regional context where they are traded. Establishing these precise connections ensures semantic clarity and helps prevent errors arising from ambiguous or incorrect associations.

APTO offers several examples that underscore the importance of this type of modeling. For example, a frequent ontological modeling error appears in sources such as Agrotermos, Agrovoc, and even in Wikipedia, concerning the term `Pimenta.' In Brazilian Portuguese, `Pimenta' is often translated into English as `pepper.' While this translation might seem correct, it oversimplifies the term and creates semantic confusion. In Brazilian Portuguese, `Pimenta' refers to the reddish or yellowish (sometimes green when collected unripe) berries from plants of the genus \textit{Capsicum}. Examples include `Pimenta malagueta' (\textit{Capsicum frutescens}), `Pimenta cumari' (\textit{Capsicum baccatum var. praetermissum}), and `Pimenta dedo-de-moça' (\textit{Capsicum baccatum}).

In resources such as Agrovoc \cite{agrovoc_pepper} and Agrotermos \cite{agrotermos_pimenta}, the concept of `Pimenta' is translated into English as `pepper' and associated with the species \textit{Piper nigrum}. In Brazil, however, \textit{Piper nigrum} refers explicitly to the plant that produces black pepper, known locally as `Pimenta-do-reino.'

In Agrotermos, \textit{Capsicum} sp. is listed as a related concept to `Pimenta,' but there is also a relationship indicating that  `Pimenta' is produced by \textit{Piper nigrum}. This misrepresentation creates confusion, as the term `Pimenta' in Brazil is predominantly associated with \textit{Capsicum} sp., while `Pimenta-do-reino' refers to \textit{Piper nigrum} (black pepper), as previously mentioned. 

Despite being a common mistake, the organisms represented by these concepts are biologically distinct. Table \ref{tab:classification_comparison} shows the biological classification of both species, highlighting their differences. \textit{Capsicum} sp. and \textit{Piper nigrum} are only related at the class level (\textit{Magnoliopsida}), underscoring their distinct evolutionary and biological characteristics.

\begin{table}[ht]
    \centering
    \caption{Classification of \textit{Piper nigrum} and \textit{Capsicum} sp., based on the GBIF Backbone Taxonomy \cite{gbif2023backbone}.}
    \label{tab:classification_comparison}
    {\scriptsize 
\renewcommand{\arraystretch}{0.9}
    \begin{tabular}{>{\raggedright}p{2.5cm} p{3.5cm} p{3.5cm}}
        \hline
        \textbf{Taxon rank} & \textbf{Piper nigrum} & \textbf{Capsicum sp.} \\
        \hline
        \textbf{Kingdom}        & Plantae               & Plantae               \\
        \textbf{Phylum}         & Tracheophyta          & Tracheophyta          \\
        \textbf{Class}          & Magnoliopsida         & Magnoliopsida         \\
        \textbf{Order}          & Piperales             & Solanales             \\
        \textbf{Family}         & Piperaceae            & Solanaceae            \\
        \textbf{Genus}          & \textit{Piper} L.     & \textit{Capsicum} L.  \\
        \textbf{Species}        & \textit{Piper nigrum} L. & ---                 \\
        \hline
    \end{tabular}
    }
\end{table}

The accurate English equivalent of the Brazilian concept `Pimenta' is `chili pepper.' However, in Agrovoc \cite{agrovoc_Chillies} and Agrotermos \cite{agrotermos_Condimento}, the concept of `chili pepper' is incorrectly matched with the Portuguese term `Condimento picante', which translates to `spicy condiment' in English. This translation conflates two distinct concepts from different hierarchical levels, treating them as semantic equivalents when they are not.

\begin{figure}[ht]
    \centering
    \begin{subfigure}{0.48\linewidth}
        \centering
        \includegraphics[width=\linewidth]{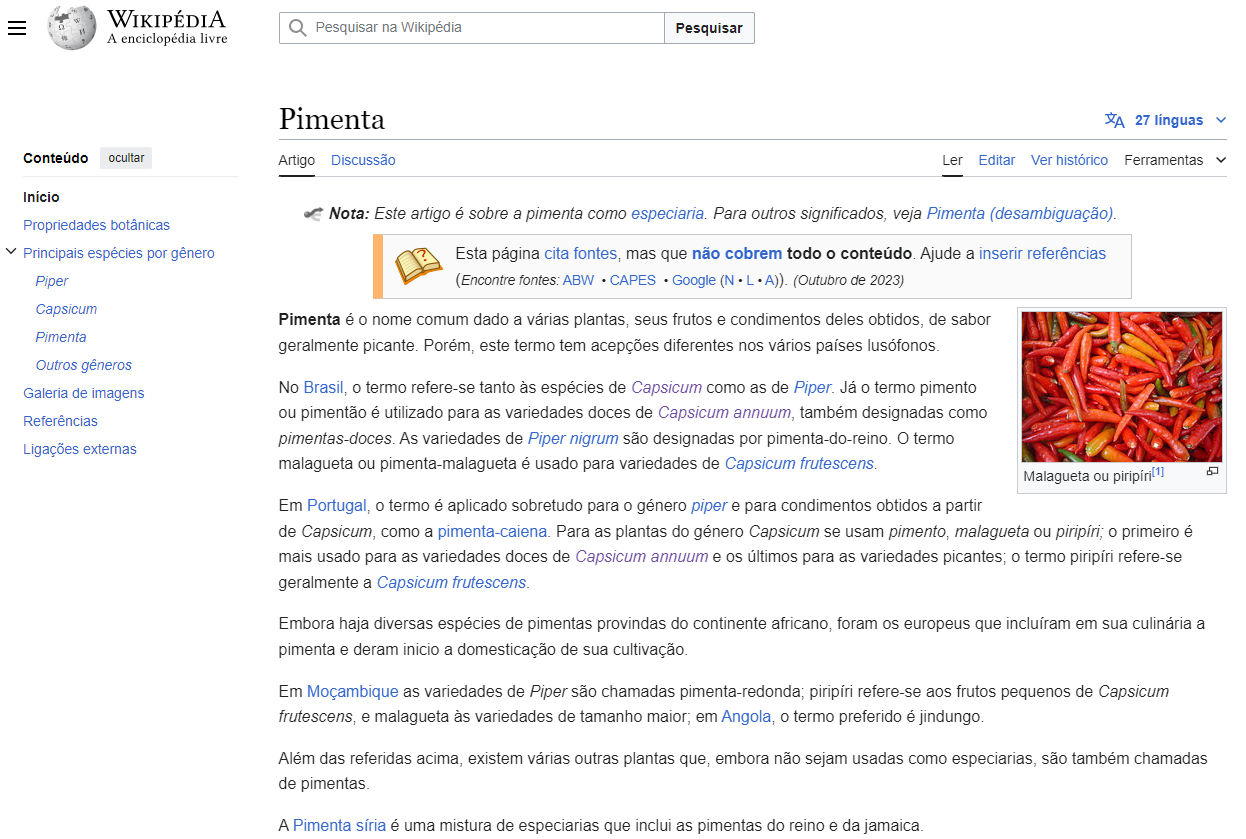}
        \caption{Pimenta on Wikipedia in Portuguese.}
        \label{fig:wiki_pimenta}
    \end{subfigure}
    \hfill
    \begin{subfigure}{0.48\linewidth}
        \centering
        \includegraphics[width=\linewidth]{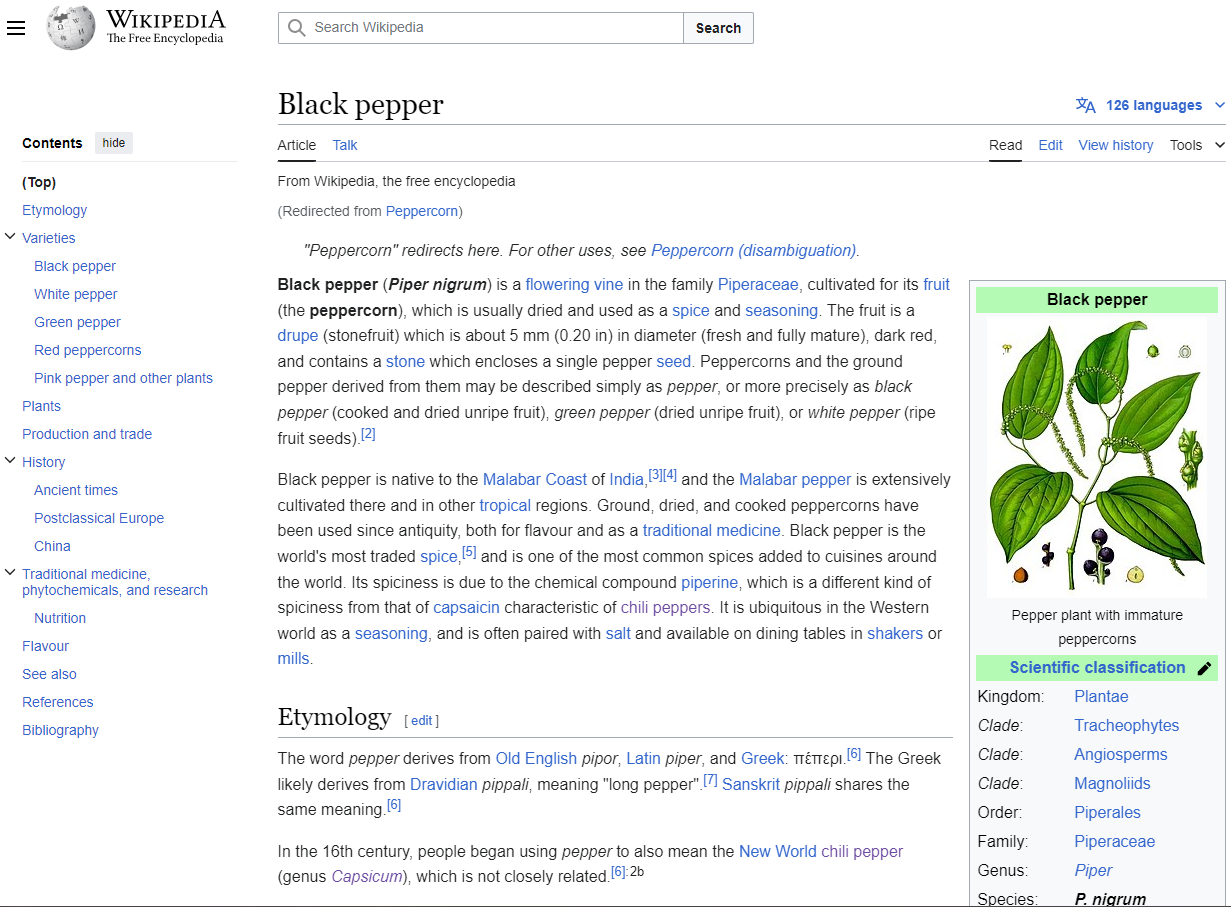}
        \caption{Black pepper as direct translation of `Pimenta' on Wikipedia in English.}
        \label{fig:wiki_piper}
    \end{subfigure}
    \caption{`Pimenta' and its translation as Black pepper in Wikipedia\protect\footnotemark.}
    \label{fig:wiki_comparison}
\end{figure}

This issue is not limited to Agrotermos and AGROVOC. On Wikipedia, a search for `Pimenta' in Portuguese retrieves an article describing `Pimentas' derived from \textit{Capsicum} sp., as shown in Fig. \ref{fig:wiki_pimenta}. However, switching the article's language to English leads to an article about `pepper,' which refers to \textit{Piper nigrum}, as illustrated in Fig. \ref{fig:wiki_piper}.

This translation is problematic because the term `Pimentas chili' is not even used in Brazil, leading to further confusion. These examples highlight how a single incorrect association can trigger a series of ontological inconsistencies across different terminology systems and languages.

\footnotetext{Wikipedia sources: \url{https://pt.wikipedia.org/wiki/Pimenta}, \url{https://en.wikipedia.org/wiki/Black_pepper}.}

To address this issue in APTO, we created two distinct classes, each explicitly linked to the correct biological classification:

\begin{itemize}
    \item \texttt{Pepper: `member of taxon' exactly 1 \textit{Piper nigrum}}
    \item \texttt{Chili pepper: `member of taxon' some \textit{Capsicum}}
\end{itemize}

Multilingual vocabularies often face ontological challenges due to the complexity of modeling concepts across diverse regions and languages. However, such issues are not exclusive to global contexts: regionalisms can also create significant ontological challenges at the local level. A prominent example is \texttt{apto:Cheiro\_verde}, a concept representing a commonly used culinary herb mixture in Brazil. The term, which literally translates to `green smell' in English, is a staple in Brazilian cuisine, enhancing the flavor of soups, stews, and meats.

\begin{figure}[ht]
\centering
\includegraphics[width=0.8\linewidth]{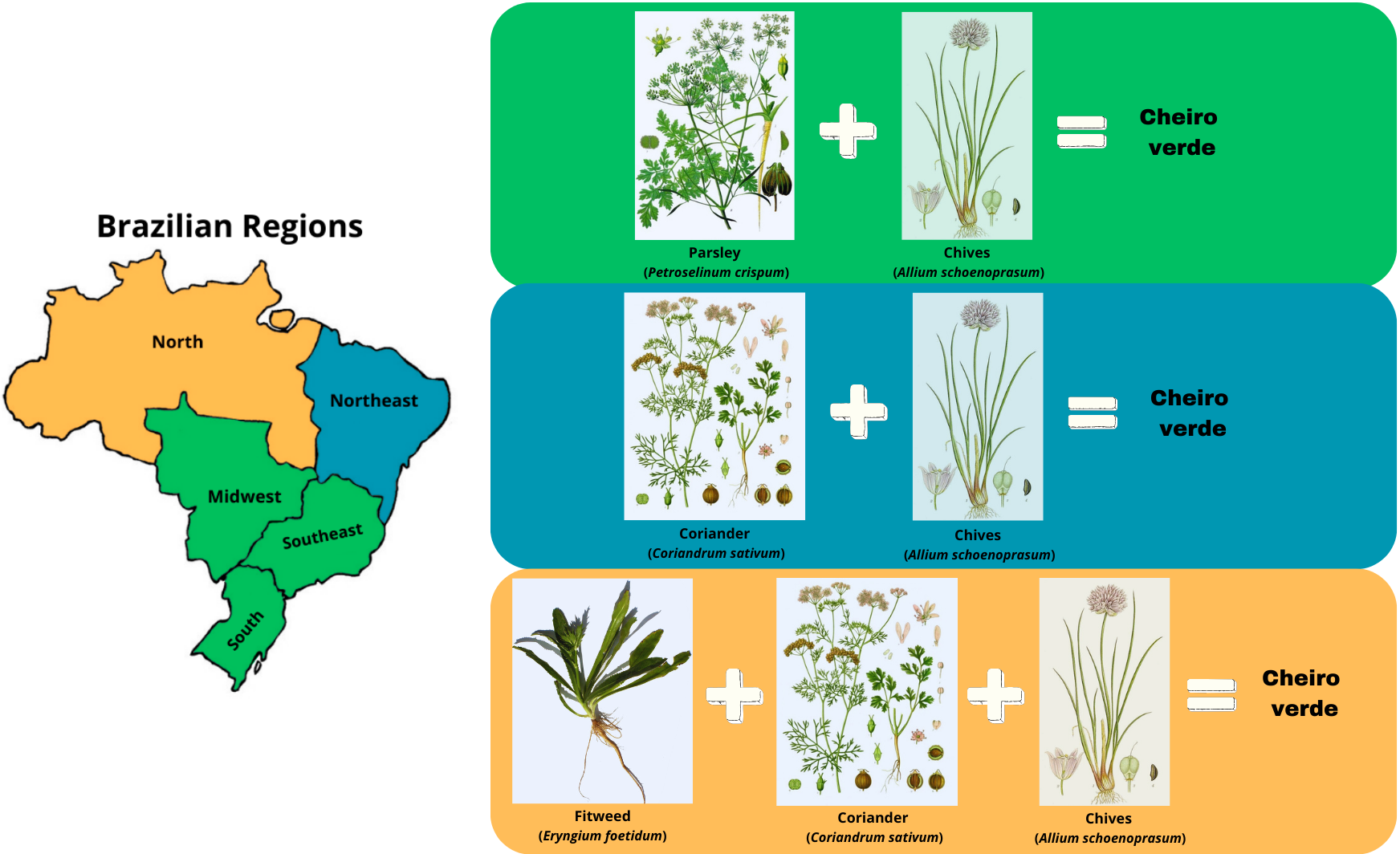}
\caption{Herbs included in `cheiro verde' by Brazilian regions. Parsley image source: \cite{kohler1897Petroselinum}; Coriander image source: \cite{kohler1897Coriandrum}; Chives image source: \cite{lindman1917allium}; Fitweed image source: \cite{Cayngotau2022}. All images reused in this figure are licensed under the public domain on Wikimedia Commons.}
\label{fig:cheiro_verde}
\end{figure}

The composition of `Cheiro verde' varies significantly across Brazil’s regions, as illustrated in Fig. \ref{fig:cheiro_verde}. In the Southeast, South, and Midwest, it typically consists of parsley and chives. In the Northeast, parsley is replaced by coriander; in the North, the mixture expands to include chives, coriander, and fitweed. This regional variation is particularly important for traditional Amazonian dishes such as `Pato no tucupi' (duck in tucupi) and `Tacacá' \cite{brasil2002alimentos,menezes2021cheiro}.

In APTO, the variations of this concept have been modeled as shown in Listing \ref{lst:cheiro}. This approach allows even non-Portuguese speakers to understand the basic semantics of the concept despite its lack of a direct English translation. The class \texttt{Cheiro\_verde} is defined as a combination of \texttt{Coriander and Chives}, \texttt{Parsley and Chives}, or \texttt{Coriander and Chives and Fitweed}, with each ingredient linked to its corresponding species. These combinations should ideally be associated with the specific Brazilian regions to refine this modeling further.

\begin{lstlisting}[language=owl, caption=`Cheiro verde' representation in APTO, label=lst:cheiro, breaklines=true, basicstyle=\footnotesize\ttfamily]
Class: `Cheiro verde'
    SubClassOf: `Culinary herbs'
    'Equivalent to'(Coriander and Chives) or 
    (Coriander and Chives and Fitweed) or 
    (Parsley and Chives)
...
Class: `Coriander'
    `member of taxon' exactly 1 `Coriandrum sativum'

Class: `Chives'
    `member of taxon' exactly 1  `Allium schoenoprasum'

Class: `Fitweed'
    `member of taxon' exactly 1 `Eryngium foetidum'

Class: `Parsley'
    `member of taxon' exactly 1 `Petroselinum crispum'    
\end{lstlisting}

\section{Final Considerations}\label{sec:final}

This study explored two approaches for integrating taxonomic data into APTO, leveraging ChatGPT with the BrowserOP plugin (Approach 1) and a Python algorithm developed with ChatGPT's assistance (Approach 2). Each approach demonstrated specific strengths and weaknesses, highlighting the potential and challenges of integrating Artificial Intelligence and automation in ontology development.

Approach 1 utilized ChatGPT for direct interaction with the GBIF API, demonstrating its capability to handle complex queries, retrieve taxonomic information, and verify synonym relationships. However, scalability limitations became evident as the dataset size increased, leading to longer processing times and issues such as non-existing URIs and incomplete tasks. The subsequent discontinuation of GPT plugins further limited the reproducibility of this approach.

In contrast, Approach 2 addressed these limitations by employing a Python algorithm to process more extensive lists of species names efficiently. This approach successfully handled a list of 74 plant species in just over 2.5 minutes while maintaining consistent class-subclass relationships and avoiding redundancy in taxonomic hierarchies. However, it was sensitive to typographical errors, requiring a pre-processing step with ChatGPT for name validation.

The study underscores the complementary strengths of generative AI and algorithmic methods in ontology development. ChatGPT proved effective for tasks such as synonym detection, taxonomic hierarchy generation, and OWL code generation. Its ability to learn from errors and refine subsequent outputs demonstrates its potential for interactive and adaptive ontology design. In contrast, the Python algorithm provided the scalability and robustness required for handling larger datasets, demonstrating the practicality of combining AI-driven tools with structured programming techniques.

Challenges remain, particularly regarding dependency on external tools like ChatGPT, which may evolve unpredictably, as seen with the discontinuation of GPT plugins. Additionally, addressing semantic variations, such as those exemplified by the concepts `Pimenta' and `Cheiro Verde,' requires contextual modeling that goes beyond simple automation. 

The workflow developed in this study is adaptable beyond the agricultural domain and can be applied to fields such as biodiversity informatics, ecological modeling, and conservation planning. Future work should explore fully automated solutions for error detection and validation, potentially incorporating machine learning models to reduce reliance on manual intervention. A complementary algorithm should be developed to periodically verify all species names in the ontology and update them in alignment with the latest version of the GBIF Backbone Taxonomy, for each new ontology version release. 

In conclusion, this paper demonstrates how combining AI-driven tools can streamline ontology engineering, address domain-specific challenges, and enhance scalability and efficiency. 

\section*{Data and Code Availability}

All data and code associated with this paper have been made openly available on Zenodo:

\begin{enumerate}
    \item Soares, F. M., Saraiva, A. M., Ferreira Pires, L., da Silva Santos, L. O. B., de Abreu Moreira, D., Corrêa, F. E., Braghetto, K. R., Pignatari Drucker, D., \& Botazzo Delbem, A. C. (2025). Supporting Data for ``Exploring ChatGPT-4 for Transforming Taxonomic Data into OWL: Lessons Learned and Implications for Ontology Development" [Data set]. Version 2. Zenodo. \url{https://doi.org/10.5281/zenodo.14982527}\label{code:12684941}
    \item Soares, F. M. (2025). Taxonomy OWLizer. [Software]. Version 2. Zenodo. \url{https://doi.org/10.5281/zenodo.14984136}\label{code:13328562}
\end{enumerate}

\section*{Author Contributions}

F.M.S. conceived and conducted the experiments and drafted the original manuscript. L.F.P., L.O.B.S.S., and A.M.S. supervised the study. F.M.S., D.A.M., F.E.C., K.R.B., D.P.D., and A.C.B.D. contributed to the formal analysis. All authors reviewed and approved the final text. 

\section*{Acknowledgements}
F.M.S. acknowledges support from the Fundação de Amparo à Pesquisa do Estado de São Paulo (FAPESP) under grant numbers 21/15125-0 and 22/08385-8. K.R.B. received funding from FAPESP
(Grant n. 2023/00779-0).  F.M.S., D.P.D., K.R.B., F.E.C., D.A.M., A.C.B.D., and A.M.S. thank the Center for Artificial Intelligence (C4AI), a partnership of USP, IBM, and FAPESP (Grant 2019/07665-4). We also thank Rubens Rangel Silva for supporting the design of Fig. \ref{fig:cheiro_verde}.

\section*{Competing Interests} The authors declare that they have no competing interests.

\section*{Declaration of AI Use in the Writing Process}

During the preparation of this work the authors used ChatGPT-4o with Scholar AI and Grammarly Education in order to review the text flow, grammar, spelling, and to eliminate redundant text. After using these tools, the authors reviewed and edited the content as needed and take full responsibility for the content of the published article.

\bibliographystyle{IEEEtran}

\renewcommand{\arraystretch}{1.2} 
\begin{table}[ht]
    \centering
    {\small 
    \begin{tabular}{m{3cm} m{10cm}} 
    
        \includegraphics[width=2.8cm]{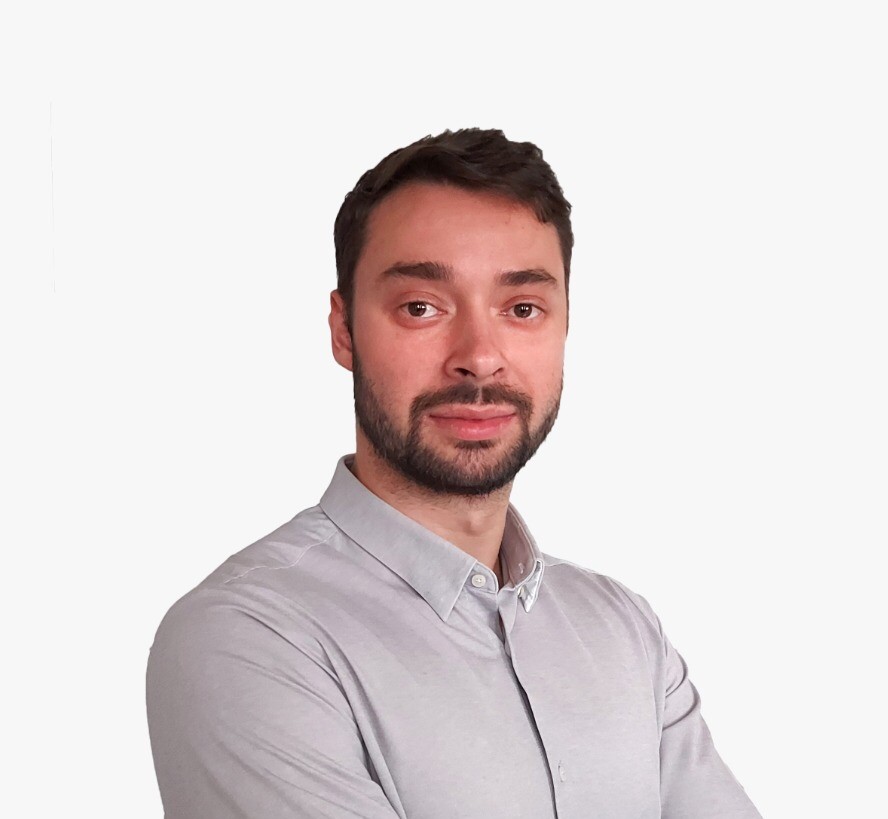} &
        \textbf{Filipi Miranda Soares}, M.Sc. (Knowledge Management), is a Ph.D. candidate in Computer Engineering at the University of São Paulo, Brazil, and in Computer Science at the University of Twente, the Netherlands, as part of a double degree program. His research focuses on integrating ontologies, Knowledge Graphs, and Generative Artificial Intelligence for enhanced agricultural data analysis.  
        \\ \multicolumn{2}{l}{\textbf{ORCID:} 0000-0002-0674-7960} \\ 
        
        \includegraphics[width=2.8cm]{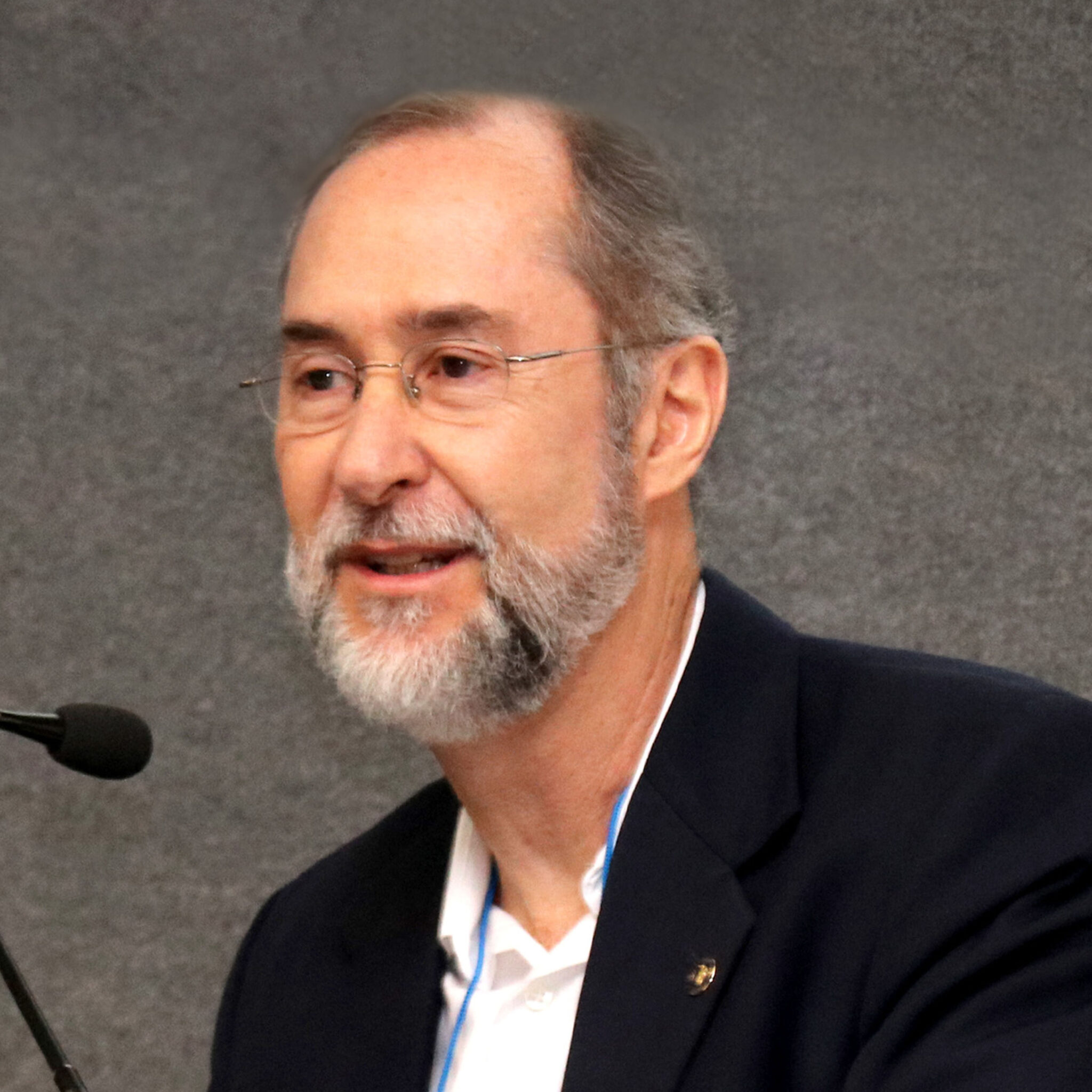} &
        \textbf{Antonio Mauro Saraiva}, Ph.D. (Electrical Engineering), is a Senior Full Professor at the Institute of Advanced Studies (IEA) at USP. From 1989 to 2024, he was a professor at the Department of Computer Engineering and Digital Systems (Polytechnic School - USP), serving as Full Professor from 2008 to 2024, until his retirement. He was Head of the Department (2014–2016), Associate Provost for Research at USP (2014–2016), and Advisor to the Research Provost (2016–2020). He chaired the Research Committees at Poli-USP (2011–2016) and IEA-USP (2018–2020).  
        \\ \multicolumn{2}{l}{\textbf{ORCID:} 0000-0003-2283-1123} \\ 
        
        \includegraphics[width=2.8cm]{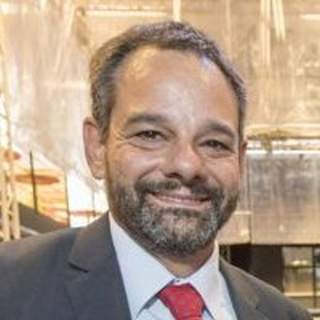} &
        \textbf{Luís Ferreira Pires}, Ph.D. obtained his M.Sc. degree from the University of São Paulo (São Paulo, Brazil) in 1989, and his PhD degree from the University of Twente (Enschede, the Netherlands) in 1994. He is currently an Associate Professor of the Faculty of Electrical Engineering, Mathematics and Computer Science of the University of Twente. His main research interests are service architectures  and platforms, design methodologies for distributed systems, and modelling and specification techniques, including Model-Driven Engineering and semantic technologies (ontologies, Knowledge Graphs and Linked Data).
        \\ \multicolumn{2}{l}{\textbf{ORCID:} 0000-0001-7432-7653} \\

        \includegraphics[width=2.8cm]{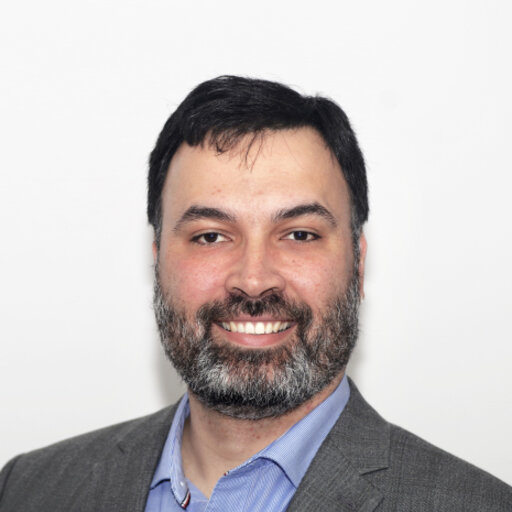} &
        \textbf{Luiz Olavo Bonino da Silva Santos}, Ph.D. He is an Associate Professor at the Semantic, Cybersecurity and Service group at the University of Twente and at the BioSemantics group, in the department of Human Genetics at the Leiden University Medical Center. His main research topics are the FAIR principles, semantic interoperability, ontology-driven conceptual modelling and intelligent systems design.
        \\ \multicolumn{2}{l}{\textbf{ORCID:} 0000-0002-1164-1351} \\        

        \includegraphics[width=2.8cm]{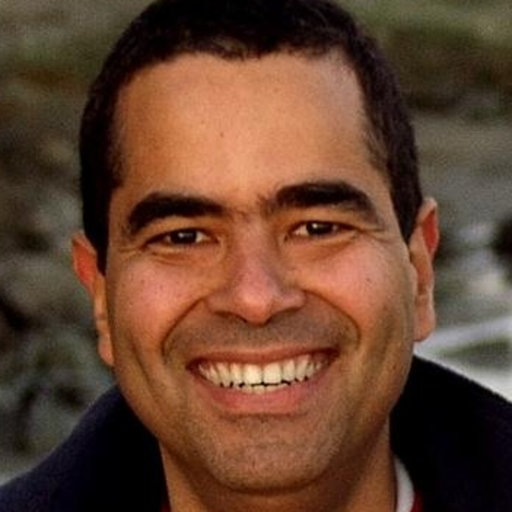} &
        \textbf{Dilvan de Abreu Moreira}, Ph.D. (Electronics Engineering). PostDoc in Biomedical Informatics from Stanford University (2008 - USA), Ph.D. in Electronics Engineering from University of Kent at Canterbury (1995 - UK), M.Sc. in Microelectronics from State University of Campinas (1991 - Brazil), B.S. on Electric Engineering from Federal University of Bahia - UFBa (1988 - Brazil). Currently Associate Professor (tenured) at University of São Paulo - USP (since 1997).
        \\ \multicolumn{2}{l}{\textbf{ORCID:} 0000-0002-4801-2225} \\    
        
    \end{tabular}
    }
\end{table}

\renewcommand{\arraystretch}{1.2} 
\begin{table}[ht]
    \centering
    {\small 
    \begin{tabular}{m{3cm} m{10cm}} 
        \includegraphics[width=2.8cm]{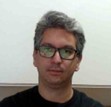} &
        \textbf{Fernando Elias Corrêa}, Ph.D. Phd. is a Data Science specialist and holds both master’s and PhD degrees in Computer Engineering from São Paulo University (USP). Currently, is a Postdoctoral Researcher in data management applied to agribusiness at the Center for Artificial Intelligence – C4AI. His research is focused on artificial intelligence, Spatial Temporal and Data management.
        \\ \multicolumn{2}{l}{\textbf{ORCID:} 0000-0002-2744-6104} \\ 
        
        \includegraphics[width=2.8cm]{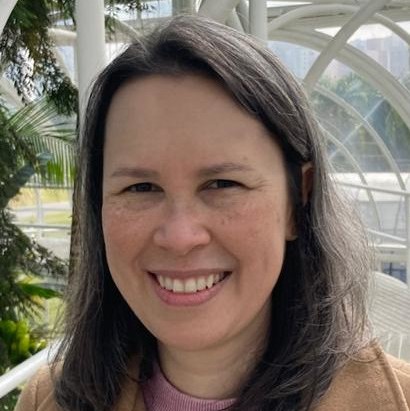} &
        \textbf{Kelly Rosa Braghetto}, Ph.D., is an Assistant Professor in the Department of Computer Science at the Institute of Mathematics and Statistics, University of São Paulo. She earned her Ph.D. (2011) and M.Sc. (2006) in Computer Science from the University of São Paulo. Her primary research interests include data engineering, data integration, scientific data management, and smart cities.
        \\ \multicolumn{2}{l}{\textbf{ORCID:} 0000-0001-6218-6849} \\ 

        \includegraphics[width=2.8cm]{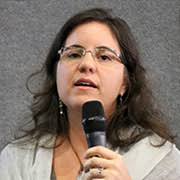} &
        \textbf{Debora Pignatari Drucker}, Ph.D. (Environment and Society) is an analyst at Embrapa Digital Agriculture, a Brazilian Agricultural Research Corporation Research Center. She coordinates the GO FAIR Brazil Agriculture Thematic Network and is one of the co-chairs of the Research Data Alliance IGAD Community of Practice (Improving Global Agricultural Data). She is also an expert at the Data and Knowledge Task Force of the Intergovernmental Science-Policy Platform on Biodiversity and Ecosystem Services (IPBES). Her major interest is in mobilizing high-quality data to support scientific research and decision-making.
        \\ \multicolumn{2}{l}{\textbf{ORCID:} 0000-0003-4177-1322} \\   
        
        \includegraphics[width=2.8cm]{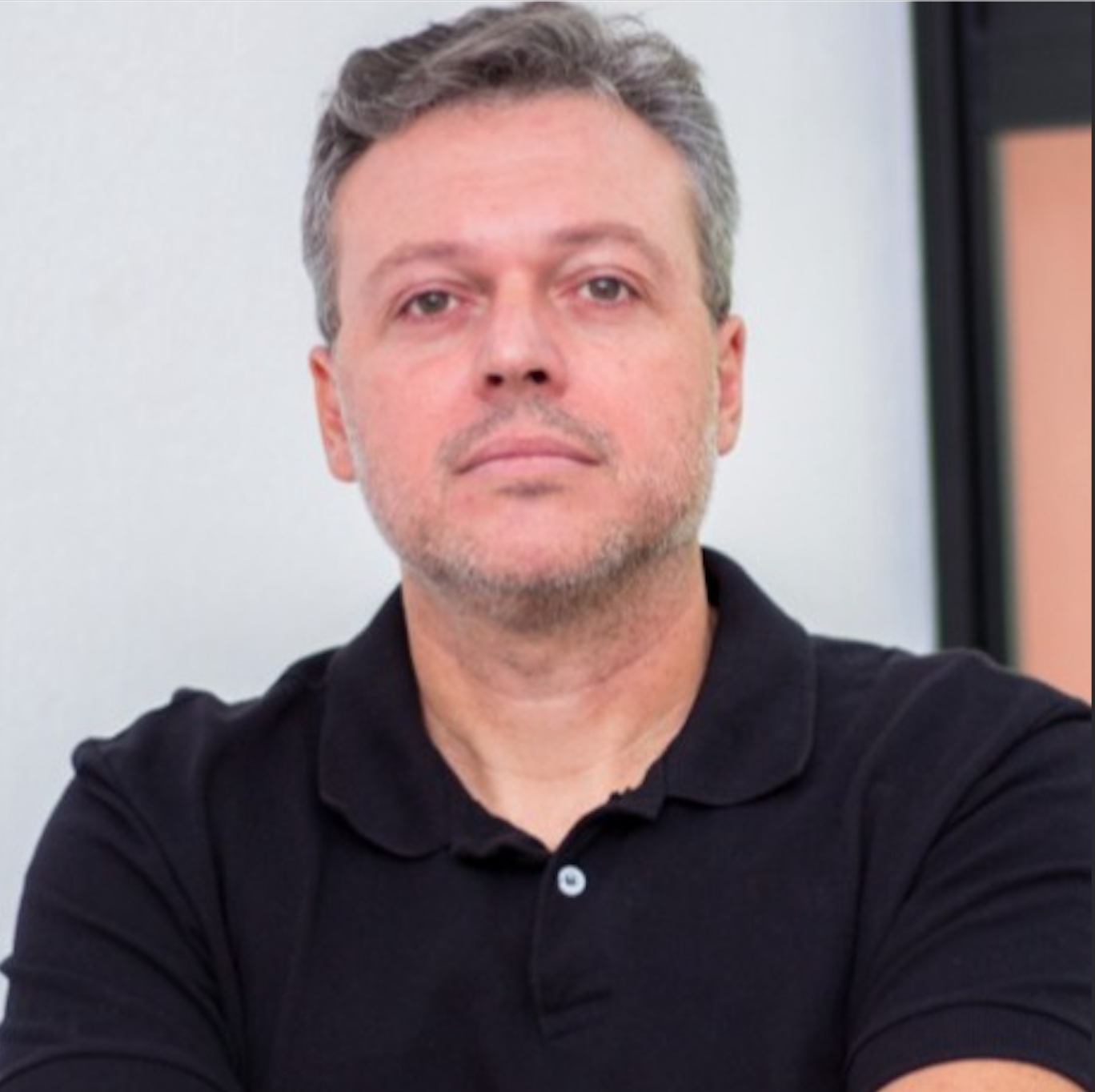} &
        \textbf{Alexandre Cláudio Botazzo Delbem}, Ph.D. Full Professor at the Department of Computer Systems of the Institute of Mathematical and Computer Sciences in the University of São Paulo (ICMC-USP). His research focuses on computational intelligence for addressing the complexity of various problems, including multiobjective combinatorial optimization, multicriteria decision-making, and explainable artificial intelligence applied to unstructured datasets. His methods have been applied across social, environmental, health, and engineering domains. 
        \\ \multicolumn{2}{l}{\textbf{ORCID:} 0000-0003-1810-1742} \\  
        
    \end{tabular}
    }
\end{table}

\end{document}